\documentclass[letterpaper]{article} 
\usepackage[]{aaai23}
\usepackage{times}  
\usepackage{helvet}  
\usepackage{courier}  
\usepackage[hyphens]{url}  
\usepackage{graphicx} 
\usepackage{natbib}  
\usepackage{caption} 
\usepackage{algorithm}
\usepackage{algorithmic}

\usepackage{newfloat}
\usepackage{listings}
\DeclareCaptionStyle{ruled}{labelfont=normalfont,labelsep=colon,strut=off} 
\floatstyle{ruled}
\newfloat{listing}{tb}{lst}{}
\floatname{listing}{Listing}
\usepackage{amsfonts}
\usepackage{xcolor} 
\newcommand{\paragraphb}[1]{\noindent{\bf #1} }
\newtheorem{theorem}{Theorem}[section]
\newtheorem{lemma}[theorem]{Lemma}
\newtheorem{corollary}[theorem]{Corollary}
\newcommand{\Name}{{FedPerm}\xspace}

\newtheorem{definition}{Definition}[section]

\definecolor{mygreen}{RGB}{77,175,74}
\definecolor{myblue}{RGB}{55,126,184}
\definecolor{skyblue}{RGB}{117,187,253}
\definecolor{myred}{RGB}{228,26,28}

\newcommand{\blue}[1]{\textcolor{blue}{#1}}

\newcommand{\clientbox}[1]{\colorbox{myblue!30}{#1}}
\newcommand{\serverbox}[1]{\colorbox{mygreen!30}{#1}}

\usepackage{xspace}
\usepackage{enumitem}
\setitemize{noitemsep,topsep=0pt,parsep=0pt,partopsep=0pt}

\usepackage{subcaption}
\usepackage{amsmath}
\usepackage{amssymb}
\usepackage{multirow}
\usepackage{comment}

\newcommand{\vjm}[1]{{\color{red} [VJM: #1]}}

\title{FedPerm: Private and Robust Federated Learning by Parameter Permutation}
\author {
    Hamid Mozaffari\textsuperscript{\rm 1}\thanks{The work was done while interning at the Oracle Labs.},
    Virendra J. Marathe\textsuperscript{\rm 2},
    Dave Dice\textsuperscript{\rm 2}
}
\affiliations {
    \textsuperscript{\rm 1} University of Massachusetts Amherst\\
    \textsuperscript{\rm 2} Oracle Labs\\
    hamid@cs.umass.edu, virendra.marathe@oracle.com, dave.dice@oracle.com
}

\usepackage{bibentry}

\begin{document}

\maketitle

\begin{abstract}

  Federated Learning (FL) is a distributed learning paradigm that
  enables mutually untrusting clients to collaboratively train a
  common machine learning model.  Client data privacy is paramount in
  FL.  At the same time, the model must be protected from poisoning
  attacks from adversarial clients.  Existing solutions address these
  two problems in isolation.  We present \Name, a new FL algorithm
  that addresses both these problems by combining a novel intra-model
  parameter shuffling technique that amplifies data privacy, with
  Private Information Retrieval (PIR) based techniques that permit
  cryptographic aggregation of clients' model updates.  The
  combination of these techniques further helps the federation server
  constrain parameter updates from clients so as to curtail effects of
  model poisoning attacks by adversarial clients.  We further present
  \Name's unique hyperparameters that can be used effectively to trade
  off computation overheads with model utility.  Our empirical
  evaluation on the MNIST dataset demonstrates \Name's effectiveness
  over existing Differential Privacy (DP) enforcement solutions in FL.
  
\end{abstract}

\section{Introduction}
Federated Learning (FL) is a distributed learning paradigm where
mutually untrusting \emph{clients} collaborate to train a shared
model, called the \emph{global model}, without explicitly sharing
their local training data. FL training involves a \emph{server} that
aggregates, using an \emph{aggregation rule} (AGR), model updates that
the clients compute using their local private data.  The aggregated
\emph{global model} is thereafter broadcasted by the server to a
subset of the clients.  This process repeats for several rounds until
convergence or a threshold number of rounds.  Though highly promising,
FL faces multiple challenges~\cite{kairouz2019advances} to its
practical deployment.  Two of these challenges are (i) data privacy
for clients' training data, and (ii) robustness of the global model in
the presence of malicious clients.

The data privacy challenge emerges from the fact that raw model
updates of federation clients are susceptible to privacy attacks by an
adversarial server as demonstrated by several recent
works~\cite{li2022,lim2021,nasr2019comprehensive,wei2020,zhu2019deep}.
Two classes of approaches can address this problem in significantly
different ways: First, \emph{Local Differential
  Privacy}~\cite{duchi13,kasiviswanathan08,truex20,warner65} in FL
(LDP-FL) enforces a strict theoretical privacy guarantee to model
updates of clients.  The guarantee is enforced by applying carefully
calibrated noise to the clients' local model updates using a local
randomizer $\mathcal{R}$.  In addition to the privacy guarantee,
LDP-FL can defend against poisoning attacks by malicious clients, thus
providing robustness to the global model
~\cite{nguyen2021flame,naseri2020local,sun2019can}.  However, the
model update perturbation needed for the LDP guarantee significantly
degrades model utility.

The other approach to enforce client data privacy is \emph{secure
  aggregation (sAGR)}, where model update aggregation is done using
cryptographic techniques such as homomorphic encryption or secure
multi-party computation~\cite{bonawitz2017practical,
  zhang2020batchcrypt, bell2020secure,fereidooni2021safelearn}.  sAGR
protects privacy of clients' data from an adversarial server because
the server sees just the encrypted version of clients' model updates.
Moreover, this privacy is enforced without compromising global model
utility.  However, the encrypted model updates themselves provide the
perfect cover for a malicious client to poison the global
model~\cite{fereidooni2021safelearn, nguyen2021flame} -- the server
cannot tell the difference between a honest model update and a
poisoned one since both are encrypted.
In this paper we answer the dual question: \emph{Can we design an
  efficient federated learning algorithm that achieves local privacy
  for participating clients at a low utility cost, while ensuring
  robustness of the global model from malicious clients?}
To that end, we present \emph{\Name}, a new FL protocol that combines
LDP~\cite{duchi13,kasiviswanathan08,warner65}, model parameter
shuffling~\cite{erlingsson2019amplification}, and \emph{computational
  Private Information Retrieval
  (cPIR)}~\cite{chor1997computationally,chang2004single,
  aguilar2016xpir, angel2018pir}
in a novel way to achieve our dual goals.

The starting point of \Name{}'s design is \emph{privacy amplification
  by shuffling}~\cite{erlingsson2019amplification}, which enables
stronger (i.e., amplified) privacy with little model perturbation
(using randomizer $\mathcal{R}$) at each client.  Crucially, our
shuffling technique fundamentally differs from prior works in that we
apply \emph{intra-model} parameter shuffling rather that the
\emph{inter-model} parameter shuffling done
previously~\cite{erlingsson2019amplification,liu2021flame,
  girgis2021shuffled}.

Next, each \Name client privately chooses its \emph{shuffling pattern}
uniformly at random for each FL round.  To aggregate the shuffled (and
perturbed) model parameters, \Name client utilizes cPIR to generate a
set of PIR queries for its shuffling pattern that allows the server to
retrieve each parameter \emph{privately} during aggregation.  All the
server observes is the shuffled parameters of the model update for
each participating client, and a series of PIR queries (i.e., the
encrypted version of the shuffling patterns).  The server can
aggregate the PIR queries and their corresponding shuffled parameters
for multiple clients to get the encrypted aggregated model.  The
aggregated model is decrypted independently at each client.

The combination of LDP at each client and intra-model parameter
shuffling achieves enough privacy amplification to let \Name\ preserve
high model utility.  At the same time, availability of the shuffled
parameters at the federation server lets it control a client's model
update contribution by enforcing norm-bounding, which is known to be
highly effective against model poisoning
attacks~\cite{nguyen2021flame,naseri2020local,sun2019can}.

Since \Name\ utilizes cPIR which relies on homomorphic encryption
(HE)~\cite{paillier1999public, damgaard2001generalisation}, it can be
computationally expensive, particularly for large models.  We present
computation/utility trade off hyper-parameters in \Name, that enables
us to achieve an interesting trade off between computational
efficiency and model utility.  In particular, we can adjust the
computation burden for a proper utility goal by altering the size and
number of shuffling patterns for the \Name clients.

We empirically evaluate \Name on the MNIST dataset to demonstrate that
it is possible to provide LDP-FL guarantees at low model utility cost.
We theoretically and numerically demonstrate a trade off between model
utility and computational efficiency. Specifically, \Name's
hyperparameters create \emph{shuffling windows} whose size can be
reduced to drastically cut computation overheads, but at the cost of
reducing model utility due to lower privacy amplification.  We
experiment with two representative shuffling window configurations in
\Name -- ``light'' and ``heavy''.  For a $(4.0, 10^{-5})$-LDP
guarantee, the light version of \Name, where client encryption, and
server aggregation needs $52.2$ seconds and $21$ minutes respectively,
results in a model that delivers $32.85\%$ test accuracy on MNIST.
The heavier version of \Name, where client encryption and server
aggregation needs $32.1$ minutes and $16.4$ hours respectively,
results in $72.38\%$ test accuracy.  Non-private FedAvg, CDP-FL and
LDP-FL provide $91.02\%$, $53.50\%$, and $13.74\%$ test accuracies for
the same $(\varepsilon, \delta)$-DP guarantee respectively.

\section{Preliminaries}

\begin{figure}[t]
\centering
\includegraphics[width=0.9\columnwidth]{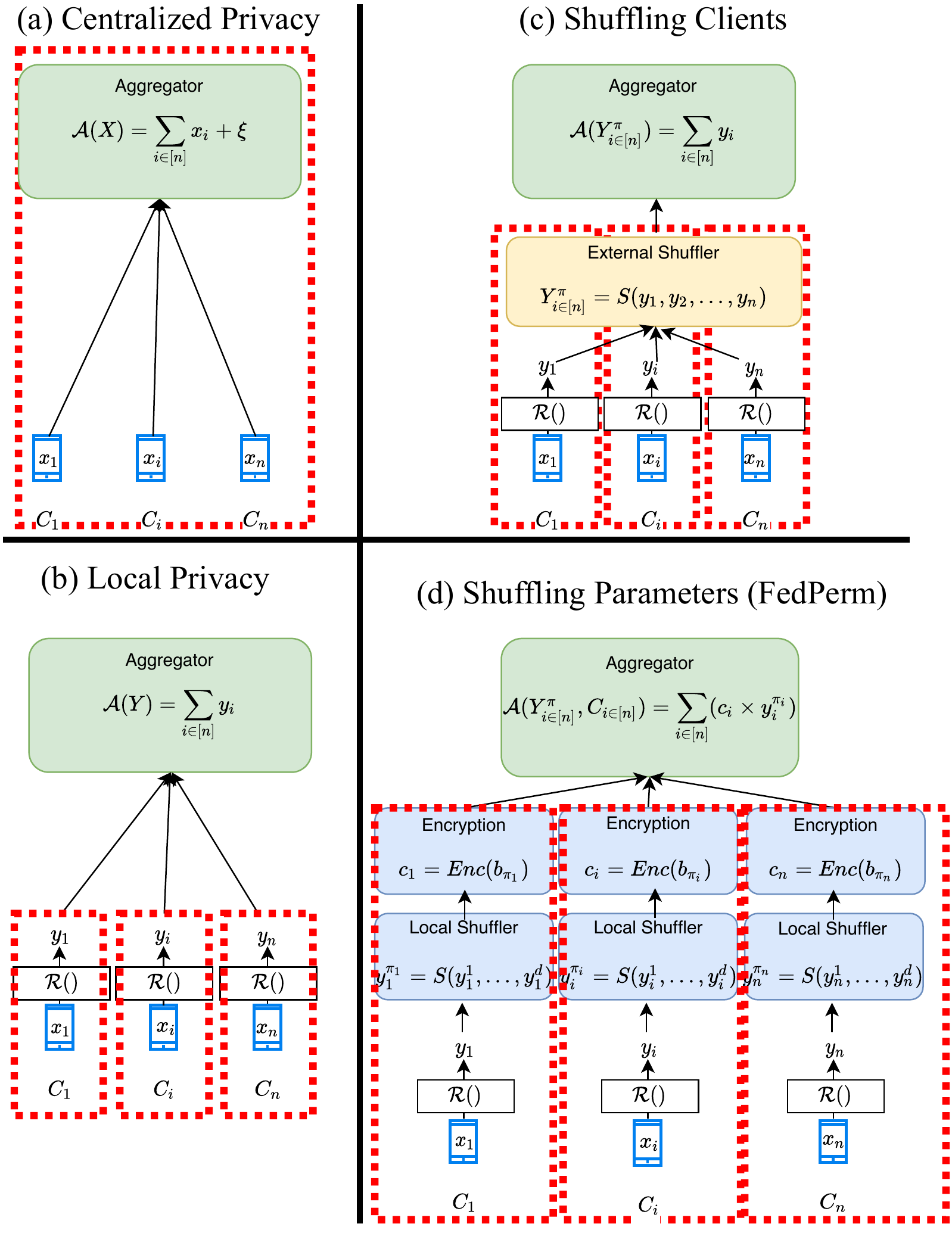} 
\vspace{-0.1in}
\caption{Different models of differential privacy in Federated
  Learning. Red dots are showing the trust boundaries.}
\label{fig:allDP}
\end{figure}

In FL~\cite{mcmahan2017communication, kairouz2019advances,
  konevcny2016federated}, $N$ clients collaborate to train a global
model without directly sharing their data.  In round $t$, the
federation server samples $n$ out of $N$ total clients and sends them
the most recent global model $\theta^t$. Each client re-trains
$\theta^t$ on its private data
using stochastic gradient descent (SGD), and sends back the model
parameter updates ($x_i$ for $i^{th}$ client) to the server. The
server then aggregates (e.g., averages) the collected parameter
updates
and updates the global model for the next round ($\theta^t \gets
\theta^{t-1} + \frac{1}{n} \sum_{i=1}^n x_i$).

\subsection{Central Differential Privacy in FL (CDP-FL)}
In CDP-FL~\cite{mcmahan2018learning, geyer2017differential},
illustrated in Figure~\ref{fig:allDP}(a) , a \emph{trusted} server
first collects all the clients' raw model updates ($x_i \in
\mathbb{R}^d$), aggregates them into the global model, and then
perturbs the model with carefully calibrated noise to enforce
differential privacy (DP) guarantees.  The server provides
participant-level DP by the perturbation.  Formally, consider
\emph{adjacent} datasets ($X, X' \in \mathbb{R}^{n\times d}$) that
differ from each other by the data of one federation client.  Then:


\begin{definition} [Centralized Differential Privacy (CDP)]
A randomized mechanism $\mathcal{M}: \mathcal{X} \rightarrow
\mathcal{Y}$ is said to be $(\varepsilon, \delta)$-differential
private if for any two adjacent datasets $X, X' \in \mathcal{X}$, and
any set $Y \subseteq \mathcal{Y}$:
\begin{equation}
\Pr [\mathcal{M}(X) \in Y] \leq e^{\varepsilon} \Pr [\mathcal{M}(X')
  \in Y] +\delta
\end{equation}
\end{definition}
where $\varepsilon$ is the privacy budget (lower the $\varepsilon$,
higher the privacy), and $\delta$ is the failure probability.

\subsection{Local Differential Privacy in FL (LDP-FL)}
CDP-FL relies on availability of a trusted server for collecting raw
model updates.  On the other hand, LDP-FL~\cite{wang2019collecting,
  liu2020fedsel} does not rely on this assumption and each client
perturbs its output locally using a randomizer $\mathcal{R}$
(Figure~\ref{fig:allDP}(b)).  If each client perturbs its model
updates locally by $\mathcal{R}$ which satisfies $(\varepsilon_{\ell},
\delta_{\ell})$-LDP, then observing collected updates
$\{\mathcal{R}(x_1), \dots, \mathcal{R}(x_n)\}$ also implies
$(\varepsilon_{\ell}, \delta_{\ell})$-DP~\cite{dwork2014algorithmic}.

\begin{definition} [Local Differential Privacy (LDP)]
A randomized mechanism $\mathcal{R}: \mathcal{X} \rightarrow
\mathcal{Y}$ is said to be $(\varepsilon_{\ell},
\delta_{\ell})$-locally differential private if for any two inputs $x,
x' \in \mathcal{X}$ and any output $y \in \mathcal{Y}$,:
\begin{equation}
\Pr [\mathcal{R}(x) = y] \leq e^{\varepsilon_{\ell}} \Pr
    [\mathcal{R}(x') = y] +\delta_{\ell}
\end{equation}
\end{definition}

In LDP-FL, each client perturbs its local update ($x_i$) with
$\epsilon_{\ell}$-LDP.  Unfortunately, LDP hurts the utility,
especially for high dimensional vectors.  Its mean estimation error is
bounded by $O(\frac{\sqrt{d \log{d}}}{\varepsilon_{\ell} \sqrt{n}})$
meaning that for better utility we should increase the privacy budget
or use larger number of users in each
round~\cite{bhowmick2018protection}.

\subsection{Privacy Amplification by Shuffling Clients' updates}
Recent works~\cite{liu2021flame, girgis2021shuffled} utilize the
privacy amplification effect by shuffling model parameters across
client model updates from participating clients to improve the LDP-FL
utility (illustrated in Figure~\ref{fig:allDP}(c)).  FL frameworks
based on shuffling clients' updates consists of three building
processes: $\mathcal{M}=\mathcal{A} \circ \mathcal{S} \circ
\mathcal{R}$.  Specifically, they introduce a shuffler $\mathcal{S}$,
which sits between the FL clients and the FL server, and it shuffles
the locally perturbed updates (by randomizer $\mathcal{R}$) before
sending them to the server for aggregation ($\mathcal{A}$).  More
specifically, given parameter index $i$, $\mathcal{S}$ randomly
shuffles the $i^{th}$ parameters of model updates received from the
$n$ participant clients.  The shuffler thus detaches the model updates
from their origin client (i.e. anonymizes them).  Previous
works~\cite{balle2019differentially, balcer2019separating,
  ghazi2019scalable} focused on shuffling one-dimensional data $x \in
X$, and corollary~\ref{coll-blanket-balle-generic} shows the privacy
amplification effect by shuffling.


\begin{corollary}~\cite{balle2019privacy}
In shuffle model, if $R$ is $\varepsilon_{\ell}$-LDP, where
$\varepsilon_{\ell} \leq \log{(n / \log{(1/\delta_c)})}/2$.  $M$
statisfies $(\varepsilon_{c}, \delta_{c})$-DP with $\varepsilon_{c}=O(
(1 \wedge \varepsilon_{\ell}) e^{\varepsilon_{\ell}}
\sqrt{\log{(1/\delta_c)}/n})$ where '$\wedge$' shows minimum function.
\label{coll-blanket-balle-generic}
\end{corollary}

From above corollary, the privacy amplification has a direct
relationship with $\sqrt{n}$ where $n$ is the number of selected
clients for aggregation, i.e., increasing the number of clients will
increase the privacy amplification.  Note that in \Name, the clients
are responsible for shuffling, and instead of shuffling the $n$
clients' updates (inter-model shuffling), each client locally shuffles
its $d$ parameters (intra-model shuffling).  In real-world settings
there is a limit on the value of $n$, so the amount of amplification
we can achieve is also limited.  However, in \Name we can see much
more amplification because we are shuffling the parameters and $n \ll
d$.

\subsection{Privacy Composition}
We use following naive and strong composition
theorems~\cite{dwork2010boosting} in this paper:

\begin{lemma}[N{\"a}ive Composition]
$\forall \varepsilon \geq 0, t \in \mathbb{N}$, the family of $\varepsilon$-DP mechanism 
satisfies $t\varepsilon$-DP under $t$-fold adaptive composition.
\label{lemma-PrivacyFold1}
\end{lemma}

\begin{lemma}[Strong Composition]
$\forall \varepsilon, \delta, \delta' > 0, t \in \mathbb{N}$, the family of $(\varepsilon, \delta)$-DP mechanism satisfies
$(\sqrt{2t \ln{(1/\delta'})} \cdot \varepsilon + t \cdot \varepsilon (e^{\varepsilon} - 1), t \delta + \delta' )$-DP
under $t$-fold adaptive composition. 
\label{lemma-PrivacyFold2}

\end{lemma}

\section{\Name: Private and Robust Federated Learning by parameter
  Permutation}

We assume a dual threat model setting where (i) the federation server
acts as an \emph{honest but curious} aggregator, and (ii) the
federation clients can maliciously attempt to poison the trained model
using manipulated local parameter updates.

\subsection{FedPerm: Design}

\Name utilizes computational Private Information Retrieval
(cPIR)~\cite{chor1997computationally,stern1998new} for secure
aggregation at the federation server.  In particular, \Name\ uses the
cPIR algorithm by Chang~\cite{chang2004single} that leverages the
algorithm by Paillier~\cite{paillier1999public}.
Algorithm~\ref{alg:algorithm1} depicts \Name.
Figure~\ref{fig:allDP}(d) depicts the \Name framework that consists of
three components, $\mathcal{F} = \mathcal{A} \circ \mathcal{S} \circ
\mathcal{R}^{d}$, denoting the client-side parameter randomizer
($\mathcal{R}^d$), the client-side shuffler ($\mathcal{S}$), and the
server-side aggregator ($\mathcal{A}$).

\paragraphb{Key Distribution} Paillier is a partial HE (PHE) algorithm
that relies on a public key encryption scheme (details of Paillier HE
in Appendix~\ref{sec:HE}).  Since Paillier is employed to protect
client updates from a curious federation server, \Name\ requires an
independent key server that generates a pair of public and secret
homomorphic keys $(Pk, Sk)$.  This key pair is distributed to all
federation clients, and just the public key $Pk$ is sent to the
federation server (for aggregation).  The key server itself can be
implemented as an independent third party server, or a leader among
the federation clients may be chosen to play that
role~\cite{zhang2020batchcrypt}.

\begin{algorithm}[h!]
\caption{\footnotesize \Name where \serverbox{green} and \clientbox{blue} colors show execution by server and client respectively.}
\label{alg:algorithm1}
{\scriptsize
\textbf{Input}: number of FL rounds $T$, number of local epochs $E$,
number of selected users in each round $n$, learning rate $\eta$,
local privacy budget $\varepsilon_d,$ number of model parameters $d$,
parameter update clipping threshold $C$\\
\textbf{Output}: $\theta_{g}^{T}$
\begin{algorithmic}[1] 
\STATE \serverbox{$\theta_g^{0} \gets$ Initialize weights}
\FOR{each iteration $t \in [T]$}
\STATE \serverbox{$U \gets$ set of $n$ randomly selected clients out of $N$ total clients}
\FOR{$u$ in $U$}
\STATE \clientbox{$\theta_{u}^t \gets \textsc{LocalUpdate}(\theta_g^{t}, \eta, E)$}
\STATE \clientbox{$\bar{\theta}_{u}^t \gets \textsc{Clip}(\theta_{u}^t, -C, C )$}
\STATE \clientbox{$\tilde{\theta}_{u}^t \gets (\bar{\theta}_{u}^t + C)/ (2C)$ }
\STATE \clientbox{$y_{u}^{t} \gets \textsc{Randomize}(\tilde{\theta}_{u}^t, \varepsilon_d)$}
\STATE \clientbox{$\pi_u \gets $ Shuffling pattern $\textsc{RandomPermutations} \in [1,d]$ }
\STATE \clientbox{$\tilde{y}_{u}^{t} \gets \textsc{Shuffle}(y_{u}^{t}, \pi_u)$}
\STATE \clientbox{$b_{u}^{t} \gets \textsc{BinaryMask}(\pi_u)$}
\STATE \clientbox{$c_{u}^{t} \gets \textsc{Enc}_{pk}(b_{u}^{t} )$}
\STATE \clientbox{Client $u$ sends $(\tilde{y}_{u}^{t}, c_{u}^{t})$ to the server}
\ENDFOR
\STATE \serverbox{norm bounding: $\tilde{y}_{u}^{t} \gets \tilde{y}_{u}^{t}\cdot \min(1, \frac{M}{{||\tilde{y}_{u}^{t}||}_2})$ for $u \in U$}
\STATE \serverbox{$\bar{z} \gets \frac{1}{n} \sum_{u \in U} \left( c_{u}^{t} \times \tilde{y}_{u}^{t} \right) $}
\STATE \clientbox{$z \gets \textsc{Dec}_{sk}(\bar{z})$}
\STATE \clientbox{normalize $z \gets C \cdot \left(2z - 1\right)$}
\STATE \clientbox{update model $\theta_g^{t+1} \gets \theta_g^{t} + z$}

\ENDFOR
\STATE \textbf{return} $\theta_g^{T}$
\end{algorithmic}
}
\end{algorithm}


\paragraphb{Client Local Training:} In the $t^{th}$ round, the server
randomly samples $n$ clients among total $N$ clients.  Each sampled
client locally retrains a copy of the global model it receives from
the server ($\theta_g^{t}$), optimizing the model using its local data
and local learning rate $\eta$ (Algorithm~\ref{alg:algorithm1}, line
5).

\paragraphb{Randomizing Update Parameters:} After computing local
updates $\theta_u^t$, client $u$ clips the update using threshold $C$
and normalizes the parameters to the range $[0,1]$
(Algorithm~\ref{alg:algorithm1}, lines 6-7).  Now the client applies
the randomizer (i.e., $\mathcal{R}^d$) on its local parameters to make
them $(\varepsilon_d)$-differentially private
(Algorithm~\ref{alg:algorithm1}, line 8).  We use the Laplacian
Mechanism as a local randomizer with privacy budget $\varepsilon_d$.

\paragraphb{Shuffling:} After clipping and perturbing the local
update, each client shuffles the parameters $y_u^t$ using the random
shuffling pattern $\pi_u$
(Algorithm~\ref{alg:algorithm1}, lines 9-10).  Shuffling amplifies the
privacy budget $\varepsilon_d$, which we discuss in
Section~\ref{sec:FedpermAdv}.

\paragraphb{Generating PIR queries:} Now the client encodes the
shuffle indices $\pi_u$ using a PIR protocol.
This process comprises two steps: first creating a binary mask of the
shuffled index, and then encrypting it using the public key of HE that
the client received in first step (Algorithm~\ref{alg:algorithm1} line
11-12).  Generally, a PIR client needs access to the $j^{th}$ record
privately from an untrusted PIR server that holds a dataset $\theta$
with $d$ records; i.e. the PIR server cannot know that the client
requested the $j^{th}$ record.  To do so, the PIR client creates a
unit vector (binary mask) $\Vec{b}_j$ of size $d$ where all the bits
are set to zero except the $j^{th}$ position being set to one:
\begin{equation}\label{eq:vecquery}
{\footnotesize
\Vec{b}_{j}=
\begin{bmatrix}
0 & 0 & \dots & 1 & \dots & 0 & 0\\
\end{bmatrix}
}
\end{equation}
If the PIR client does not care about privacy, it would send
$\Vec{b}_j$ to the PIR server, and the server would generate the
client's response by multiplying the binary mask into the database
matrix $\theta$ ($ \theta_{j} = \Vec{b}_{j} \times \theta $).  A PIR
technique allows the client to obtain this response without revealing
$\Vec{b}_j$ to the PIR server.  For example in~\cite{chang2004single},
the PIR client uses HE to encrypt $\Vec{b}_j$ element by element
before sending it to the PIR server.  During the data recovery phase,
the client extracts its target record by decrypting the component of
$\textsc{Enc}(\Vec{b}_j) \times \theta$.  Equation~\ref{eq:PIR11}
shows retrieving the $j^{th}$ record by this PIR query.  Note that a
HE system has a property that $m_1 \times m_2 \gets \textsc{Dec}
\left( \textsc{Enc}[m_1] \times m_2 \right)$.
\begin{equation}~\label{eq:PIR11}
{\footnotesize
\begin{aligned}
&\textsc{Dec} ( \textsc{Enc}(\Vec{b}_j) \times \theta )= \\
&\textsc{Dec} \left( \textsc{Enc}[0] \cdot \theta_1 + \dots + \textsc{Enc}[1] \cdot \theta_j + \dots + \textsc{Enc}[0] \cdot \theta_d \right) =\\
&\textsc{Dec} \left( \textsc{Enc}[\theta_j] \right) = \theta_j
\end{aligned}
}
\end{equation}
A \Name client creates $d$ PIR queries to retrieve each parameter
privately.  (In Section~\ref{sec:FedpermAdv}, we discuss additional
parameters to reduce the number of PIR queries.)  In this case, the
shuffled parameters ($\tilde{y}_{u}^{t}$) are the dataset located at
the PIR server and each shuffled index in $\pi_u$ is the secret record
row number (i.e. $j^{th}$ in above) that the PIR client is interested
in.  Client $u$ first creates $b_u^t$ which is a collection of $d$
binary masks of shuffled indices in $\pi_u$, similar to PIR query
$\Vec{b}_j$ in Equation~\ref{eq:vecquery}.  Then the client encrypts
the binary masks and sends the shuffled parameters and the PIR query
(encrypted binary masks) to the server for aggregation.

\paragraphb{Correctness:} Note that for every client $u$ and every round
$t$, decrypting the multiplication of the encrypted binary masks to
the shuffled parameters produces the original unshuffled parameters.
It means that for $y_u^t = \textsc{Dec} \left( c_{u}^{t} \times
\tilde{y}_{u}^{t} \right)$. So for any $( \tilde{y},c, )$ we have:
\begin{equation}\label{eq:PIRcorrect1}
{\footnotesize
\begin{aligned}
&\textsc{Dec} \left( c \times \tilde{y} \right)= \\
&\textsc{Dec} \left(
\begin{bmatrix}
\textsc{Enc}(\Vec{b}_{\pi_1}) \\ \textsc{Enc}(\Vec{b}_{\pi_2}) \\ \dots \\ \textsc{Enc}(\Vec{b}_{\pi_d}) \\
\end{bmatrix} \times 
\begin{bmatrix}
\tilde{y}_1 \\ \tilde{y}_2 \\ \dots \\ \tilde{y}_d \\
\end{bmatrix} 
\right) =\\
&\textsc{Dec} \left(
\begin{bmatrix}
\textsc{Enc}[0] & \dots & \textsc{Enc}[1] & \dots & \textsc{Enc}[0] \\ 
\textsc{Enc}[0] & \dots & \textsc{Enc}[1] & \dots & \textsc{Enc}[0] \\ 
\dots & \dots & \dots & \dots & \dots\\ 
\textsc{Enc}[0] & \dots & \textsc{Enc}[1] & \dots & \textsc{Enc}[0] \\
\end{bmatrix} \times 
\begin{bmatrix}
y_1^{\pi} \\ y_2^{\pi} \\ \dots \\ y_d^{\pi} \\ 
\end{bmatrix}
\right) =\\
&\textsc{Dec} \left(\begin{bmatrix}
\textsc{Enc}[y_1] & \textsc{Enc}[y_2] & \dots & \textsc{Enc}[y_d] \\
\end{bmatrix} \right) = \\
& \begin{bmatrix}
y_1 & y_2 & \dots & y_d \\
\end{bmatrix}
\end{aligned}
}
\end{equation}

\paragraphb{Server: norm bounding}
After collecting all the local updates $(\tilde{y}_{u}^{t},
c_{u}^{t})$ for selected clients in round $t$, the \Name server first
applies $\ell_2$-norm bounding to the threshold $M$ on the shuffled
parameters $\tilde{y}_{u}^{t}$ (Algorithm~\ref{alg:algorithm1}, line
15).  Note that unlike other robust AGRs, \emph{norm bounding} is the
only robust AGR scheme that does not require the true position of the
parameters because it works by calculating the $\ell_2$ norm of the
parameter updates as a whole irrespective of their order
(i.e. $\ell_2(\tilde{y}_{u}^{t}) = \ell_2(y_{u}^{t})$).  Prior
works~\cite{nguyen2021flame,naseri2020local,sun2019can} have shown the
effectiveness of norm bounding in defense against poisoning attacks by
malicious clients.

\paragraphb{Server: Aggregation}
Then the server aggregates all the updates into global update
$\bar{z}$ (Algorithm~\ref{alg:algorithm1}, line 16).  This aggregation
is averaging the update parameters for $n$ collected updates by
calculating $\frac{1}{n} \sum_{u \in U} \left( c_{u}^{t} \times
\tilde{y}_{u}^{t} \right)$.  The expression $c_{u}^{t} \times
\tilde{y}_{u}^{t}$ has the effect of ``unshuffling'' client $u$'s
parameters.  At the same time, the resulting vector is encrypted, thus
kept hidden from the server.

\paragraphb{Correctness of Aggregation:} In
Equation~\ref{eq:PIRcorrect1}, we show that $\forall t \in [T], u \in
U \;\; y_u^t = \textsc{Dec} \left( c_{u}^{t} \times \tilde{y}_{u}^{t}
\right)$.  
Based on the two main properties of a HE system 
\textbf{(a)} $m_1 \times m_2 \gets \textsc{Dec} \left( \textsc{Enc}[m_1] \times m_2 \right)$, 
\textbf{(b)} $m_1 + m_2 \gets \textsc{Dec} \left( \textsc{Enc}[m_1] + \textsc{Enc}[m_2] \right)$, and 
Equation~\ref{eq:PIRcorrect1}, we can derive the Equation~\ref{eq:PIRcorrect3}:
%

\begin{equation}\label{eq:PIRcorrect3}
{\footnotesize
  \textsc{Dec} \left( \frac{1}{n} \sum_{u \in U} ( c_{u}^{t} \times \tilde{y}_{u}^{t}) \right)=\frac{1}{n} \sum_{u \in U} y_u^t
}
\end{equation}

\paragraphb{Updating Global Model} The server aggregates local updates
$(\tilde{y}_{u}^{t}, c_{u}^{t})$ without knowing the true position of
the parameters as they are detached from their positions.  Result of
aggregation $\frac{1}{n} \sum_{u \in U} \left( c_{u}^{t} \times
\tilde{y}_{u}^{t} \right)$ is vector of encrypted parameters, and they
need to be decrypted to be used for updating the global model
(Algorithm~\ref{alg:algorithm1} lines 17-19).  This decryption is done
at each client using Paillier's secret key.

\section{Computation/Communication and Utility Tradeoff in \Name}
\label{sec:FedpermAdv}


Each \Name client perturbs its local update (vector $x_i$ containing
$d$ parameters) with randomizer $\mathcal{R}^d$ which is
$\varepsilon_d$-LDP, and then shuffles its parameters.  We use the
Laplacian mechanism as the randomizer.  Based on the n\"aive
composition theorem from Lemma~\ref{lemma-PrivacyFold1}, the
client perturbs each parameter value with $\mathcal{R}$ which
satisfies $\varepsilon_{wd}$-LDP where
$\varepsilon_{wd}=\frac{\varepsilon_d}{d}$ (Appendix~\ref{sec:Laplace}
contains additional details).  Corollary~\ref{coll-FedPermDP11} shows
the privacy amplification from $\varepsilon_d$-LDP to
$(\varepsilon_{\ell}, \delta_{\ell})$-DP after the parameter
shuffling.  Corollary~\ref{coll-FedPermDP11} is derived from
Corollary~\ref{coll-blanket-balle-generic}, by substituting the number
of participating clients $n$ by the number of parameters $d$ in the
model.

\begin{corollary}
If $\mathcal{R}$ is $\varepsilon_{wd}$-LDP, where $\varepsilon_{wd} \leq
\log{(d / \log{(1/\delta_{\ell})})}/2$, \Name $\mathcal{F}=\mathcal{A}
\circ \mathcal{S}_{d} \circ \mathcal{R}^d$ satisfies
$(\varepsilon_{\ell}, \delta_{\ell})$-DP with $\varepsilon_{\ell}=O(
(1 \wedge \varepsilon_{wd}) e^{\varepsilon_{wd}}
\sqrt{\log{(1/\delta_{\ell})}/d})$.
\label{coll-FedPermDP11}
\end{corollary}

Thus, larger the number of parameters in the model, greater is the
privacy amplification.  With large models containing millions or
billions of parameters, the privacy amplification can be immense.
However, the model dimensionality also affects the computation (and
communication) cost in \Name.  Each \Name client generates a
$d$-dimensional PIR query for every parameter in the model, resulting
in a PIR query matrix containing $d^2$ entries.  This results in a
quadratic increase in client encryption time, server aggregation time,
and client-server communication bandwidth consumption.  This increase
in communication, and more importantly computation, resources is
simply infeasible for large models containing billions of parameters.
To address this problem \Name introduces additional hyperparameters
that present an interesting trade off between
computation/communication overheads and model utility.

\subsection{\Name with Smaller Shuffling Pattern}


Instead of shuffling all the $d$ parameters, the \Name client can
partition its parameters into several identically sized windows, and
shuffle the parameters in each window with the \emph{same} shuffling
pattern.  Thus, instead of creating a very large random
shuffling pattern $\pi$ with $d$ indices (i.e.,
$\pi=\textsc{RandomPermutations}[1,d]$), each client creates a
shuffling pattern with $k_1$ indices (i.e.,
$\pi=\textsc{RandomPermutations}[1,k_1]$), and shuffles
($\mathcal{S}_{k_1}$) each window with these random indices.

The window size $k_1$ is a new \Name hyperparameter that can be used
to control the computation/communication and model utility trade off.
Once we set the size of shuffling pattern to $k_1$, each client needs
to perform $d \cdot k_1$ encryptions and consumes $O(d \cdot k_1)$
network bandwidth to send its PIR queries to the server.

\paragraphb{Superwindow:} A shuffling window size of $k_1$, partitions
each \Name client $u$'s local update $x_u$ ($d$ parameters) into
$w=d/k_1$ windows, each containing $k_1$ parameters.  Each \Name
client, independently from other \Name clients, chooses its shuffling
pattern $\pi$ uniformly at random with indices $\in [1,k_1]$, and
shuffles each window with this pattern.  This means that every
position $j$ ($1\le j \le k_1$) in each window $k$ ($1\le k \le w$)
will have the same permutation index ($\pi_j$).  Thus all of the
$j^{th}$ positioned parameters ($x_u^{(k,j)} \; \text{for} \; 1\le k
\le w$) will contain the value from the $\pi_j^{th}$ slot in window
$k$.
For a given index $j$ ($1\le j \le k_1$), we define a
\emph{superwindow} as the set of all of the parameters $x_u^{(k,j)}$
for $1\le k \le w$.  If we structure the parameter vector $x_u$ (with
$d$ parameters) as $\mathbb{R}^{k_1 \times w}$ (a matrix with $k_1$
rows and $w$ columns), each row of this matrix is a superwindow.


\begin{figure}[h]
\centering
\includegraphics[width=0.9\columnwidth]{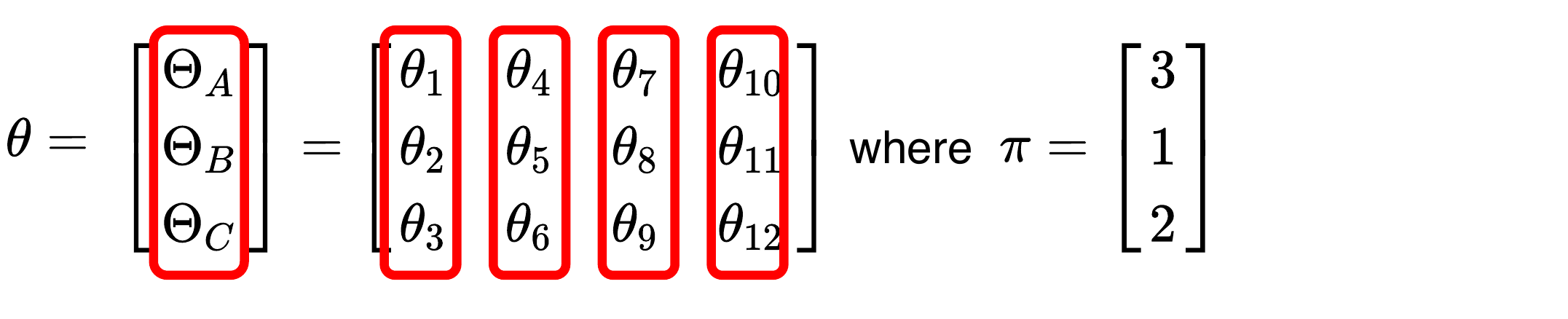} 
\vspace{-0.15in}
\caption{\Name example with $k_1=3$ and $d=12$. The red boxes are
  showing the windows that the parameters inside them are going to be
  shuffled with the same shuffling pattern $\pi$.}
\label{fig:example}
\end{figure}

Figure~\ref{fig:example} depicts an example model containing $12$
parameters $\theta=[\theta_1, \theta_2, ..., \theta_{12}]$.  The
original \Name algorithm mandates a shuffling pattern $\pi$ with $12$
indices $\in [1,12]$, where the PIR query generates $12\times12=144$
encryptions.  However, a shuffling pattern $\pi$ of three indices
$k_1=3$ ($\pi=[3,1,2]$ in the figure) requires only $3\times3=9$
encryptions.  This shuffling pattern creates $4$ windows of size $3$
(red boxes in the 2-D matrix in the figure), and each row in the 2-D
matrix, represented more succintly by $[\Theta_A,\Theta_B,\Theta_C]$,
itself constitutes a superwindow.  The shuffling pattern $\pi=[3,1,2]$
applied to $\theta = [\Theta_A,\Theta_B,\Theta_C]$ swaps entire
superwindows to give $\mathcal{S}_{k_1}(\theta) =
[\Theta_C,\Theta_A,\Theta_B]$.

Shuffling of superwindows, instead of individual parameters, leads to
a significant reduction in the computation (and communication)
overheads for \Name clients.  However, this comes at the cost of
smaller privacy amplification.  Corollary~\ref{thm-FedPermDP2a} shows
the privacy amplification of \Name from $\varepsilon_d$-LDP to
$(\varepsilon_{\ell}, \delta_{\ell})$-DP after superwindow shuffling
(with window size $k_1$).  After applying the randomizer $\mathcal{R}$
that is $\varepsilon_d$-LDP on the local parameters, each superwindow
is $\varepsilon_w$-LDP where $\varepsilon_w=w \cdot
\varepsilon_{wd}=\frac{d}{k_1} \cdot
\varepsilon_{wd}=\frac{\varepsilon_{d}}{k_1}$.  Since we are shuffling
the superwindows, we can derive Corollary~\ref{thm-FedPermDP2a} for
\Name by setting the shuffling pattern size to $k_1$ from
Corollary~\ref{coll-blanket-balle-generic}.  




\begin{corollary}
For \Name $\mathcal{F}=\mathcal{A} \circ \mathcal{S}_{k_1} \circ
\mathcal{R}^{w}$ with window size $k_1$, where $\mathcal{R}^w$ is
$\varepsilon_w$-LDP and $\varepsilon_{w} \leq \log{(k_1 /
  \log{(1/\delta_{\ell})})}/2$, the amplified privacy is
$\varepsilon_{\ell}=O( (1 \wedge \varepsilon_{w}) e^{\varepsilon_{w}}
\sqrt{\log{(1/\delta_{\ell})}/k_1}$.
\label{thm-FedPermDP2a}
\end{corollary}



\subsection{\Name with Multiple Shuffling Patterns}

An additional way to adjust the computation/communication vs. utility
trade off is by using multiple shuffling patterns.  Each \Name client
chooses $k_2$ shuffling patterns $\{\pi_1, \dots, \pi_{k_2}\}$
uniformly at random where each
$\pi_i=\textsc{RandomPermutations}[1,k_1] \; \text{for} \; 1 \le i \le
k_2$.  Then, each \Name client partitions the $d$ parameters into
$d/k_1$ windows, where it permutes the parameters of window $k$ ($1
\le k \le d/k_1$) with shuffling pattern $\pi_{i} \; \text{s.t.}\; i=k
\; \text{mod} \; k_2$.  In this case, each \Name client needs $k_2
\cdot k_1^2$ encryptions to generate the PIR queries.

\begin{figure}[h]
\centering
\includegraphics[width=0.9\columnwidth]{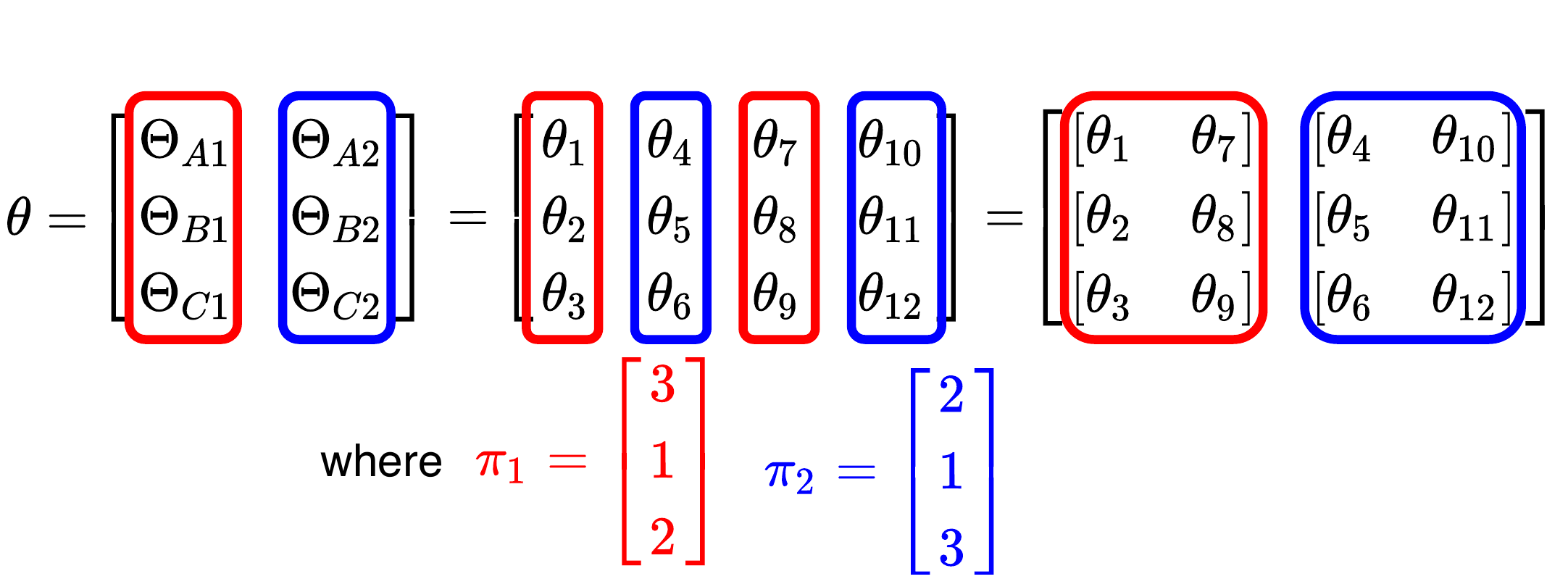} 
\vspace{-0.15in}
\caption{\Name example with $k_1=3$ and $k_2=2$. We have two shuffling patterns $\pi_1$ and $\pi_2$ shown with red and blue boxes.}
\label{fig:example2}
\end{figure}

Figure~\ref{fig:example2} shows \Name for $d=12$, $k_1=3$ and $k_2=2$,
i.e., there are two shuffling patterns $\pi_1$ (shown with red box)
and $\pi_2$ (shown with blue box) and each one has 3 shuffling
indices.  In this example, the client partitions the 12 parameters
into $4$ windows that it shuffles with $\pi_1$ (1st and 3rd windows)
and $\pi_2$ (2nd and 4th windows).  This example is equivalent to an
FL scneario with \emph{two} external inter-model shufflers (with
shuffling patterns $\pi_1, \pi_2$) and three FL clients ($A,B,C$).
Each client sends $2$ ($w=d/(k_1 k_2)$) parameters to each shuffler
for shuffling with other clients.  Two different shuffling patterns
$\pi_1$ and $\pi_2$ are applied on $[\Theta_{A1}, \Theta_{B1},
  \Theta_{C1}]$ and $[\Theta_{A2}, \Theta_{B2}, \Theta_{C2}]$
respectively.

When we have $k_2$ shuffling patterns and each shuffling pattern has
$k_1$ indices, the size of each superwindow is $w=d/(k_1 k_2)$.
Therefore, each client perturbs each superwindow with a randomizer
$\mathcal{R}^w$ that satisfies $\varepsilon_{w}$-LDP where
$\varepsilon_w=w \cdot \varepsilon_{wd}=\frac{d}{k_1 k_2} \cdot
\varepsilon_{wd}=\frac{\varepsilon_{d}}{k_1 k_2}$.  Take
$\varepsilon_w$ to Corollary~\ref{coll-blanket-balle-generic} on the
superwindows to find the amplified local privacy and then using strong
composition in Lemma~\ref{lemma-PrivacyFold2} we can easily derive the
Theorem~\ref{thm-FedPermDP3aa} for \Name with
$\mathcal{S}_{k_1}^{k_2}$.

\begin{theorem}
For \Name $\mathcal{F}=\mathcal{A} \circ \mathcal{S}_{k_1}^{k_2} \circ
\mathcal{R}^{w}$ with window size $k_1$, and $k_2$ shuffling patterns,
where $\mathcal{R}^w$ is $\varepsilon_w$-LDP and $\varepsilon_{w} \leq
\log{( k_1 / \log{((k_2 +1)/\delta_{\ell})})}/2$, the amplified
privacy is $\varepsilon_{\ell}=O( (1 \wedge \varepsilon_{w})
e^{\varepsilon_{w}} \log{(k_2/\delta_{\ell})} \sqrt{k_2/k_1})$.
\label{thm-FedPermDP3aa}
\end{theorem}


\begin{figure}[h!]
\centering
\begin{subfigure}{.22\textwidth}
\centering
\includegraphics[width=.99\linewidth]{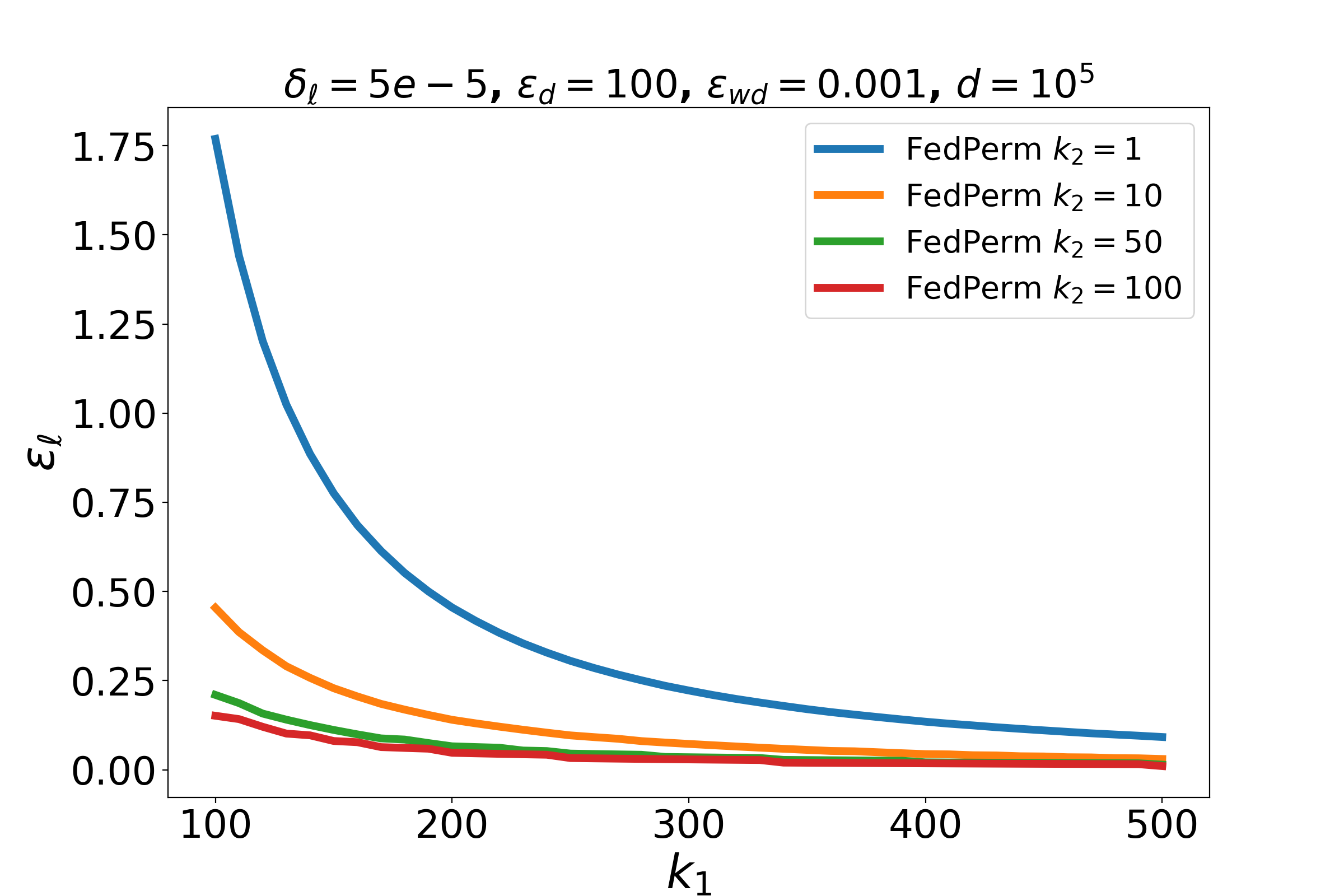} 
\caption{Impace of $k_1$ and $k_2$,}
\label{fig:Priv2}
\end{subfigure}
\begin{subfigure}{.22\textwidth}
\centering
\includegraphics[width=.99\linewidth]{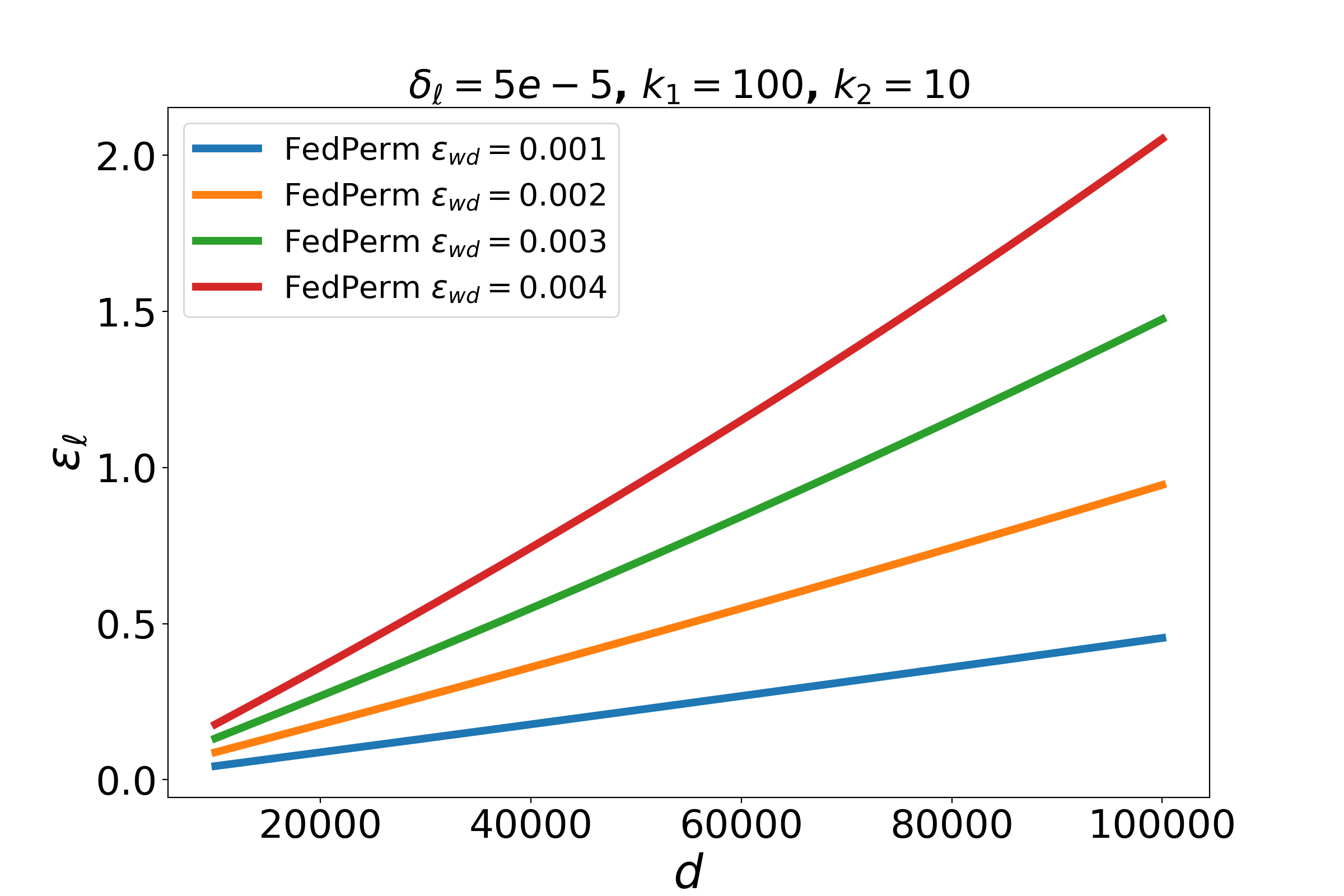} 
\caption{Impact of $d$ and $\varepsilon_{wd}$.}
\label{fig:Priv3}
\end{subfigure}
\vspace{-0.1in}
\caption{Privacy amplification of \Name from $\varepsilon_d$-LDP to
  ($\varepsilon_{\ell},\delta_{\ell}$)-DP.  We illustrate the overal
  amplification with Bennett inequality for the Laplace
  Mechanism. 
}
\label{fig:Priv-ALL}
\end{figure}

\subsection{Privacy Analysis} 
In Figure~\ref{fig:Priv-ALL}, we show the relationship of our introduced varibales $k_1$, $k_2$, $\varepsilon_d$ and $d$ on the privacy amplification in \Name. 
Figure~\ref{fig:Priv2} shows the privacy amplification effect from $\varepsilon_{d}$-LDP to ($\varepsilon_{\ell},\delta_{\ell}$)-DP for the local model updates after shuffling with $k_2$ shuffling patterns each with size of $k_1$. 
We can see that each client can use larger shuffling patterns (i.e. , increasing $k_1$) or more shuffling patterns (i.e., increasing $k_2$) and get larger privacy amplification.
However, this comes with a price where this imposes more computation/communication burden on the clients to create the PIR queries as they need to encrypt $k_2 \times k_1^2$ values 
and send them to the server, and it also imposes higher computation on the server as it should multiply larger matrices.
Figure~\ref{fig:Priv3} shows the amplification of privacy for fixed value of $k_1=100, k_2=10$ for various model sizes. 
From this figure we can see that if we want to provide same privacy level for larger models, we need to increase values of $k_1$ or $k_2$ (i.e. more computation/communication cost).

\section{Experiments}
In this section, we investigate the utility and computation trade offs
in \Name.  We use MNIST dataset and a logistic regression model with
$d=7850$ parameters to evaluate these trade offs.

\subsection{Comparison with Baselines}
We compare our results with following baselines: \textbf{(a)}
FedAvg~\cite{mcmahan2017communication} with no privacy, \textbf{(b)}
CDP-FL~\cite{mcmahan2018learning, geyer2017differential}, \textbf{(c)}
LDP-FL~\cite{wang2019collecting, liu2020fedsel} with Gaussian
Mechanism.

\begin{figure}[h!]
\centering
\begin{subfigure}{.23\textwidth}
\centering
\includegraphics[width=.99\linewidth]{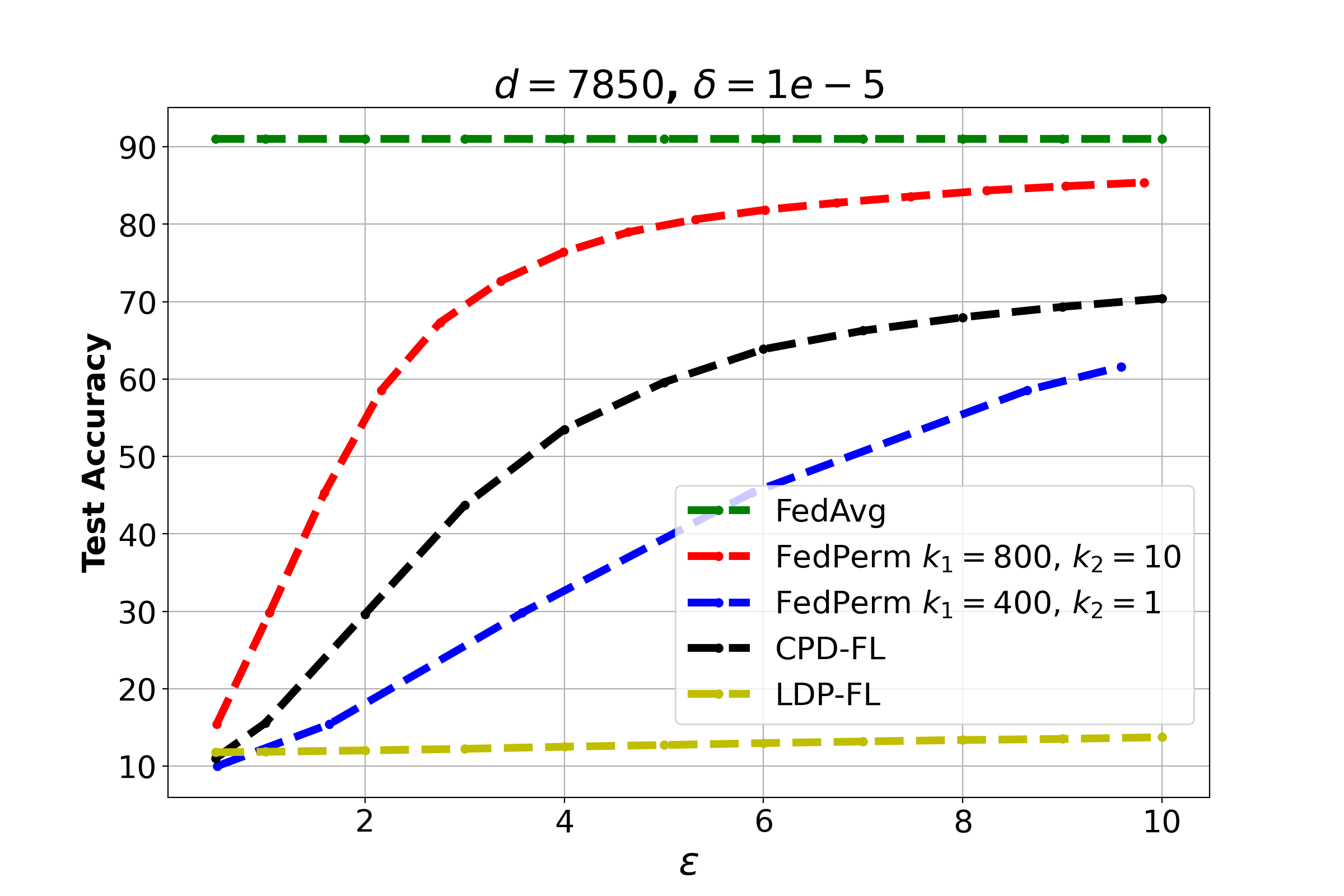} 
\caption{Final test accuracy.}
\label{fig:Acc1}
\end{subfigure}
\begin{subfigure}{.23\textwidth}
\centering
\includegraphics[width=.99\linewidth]{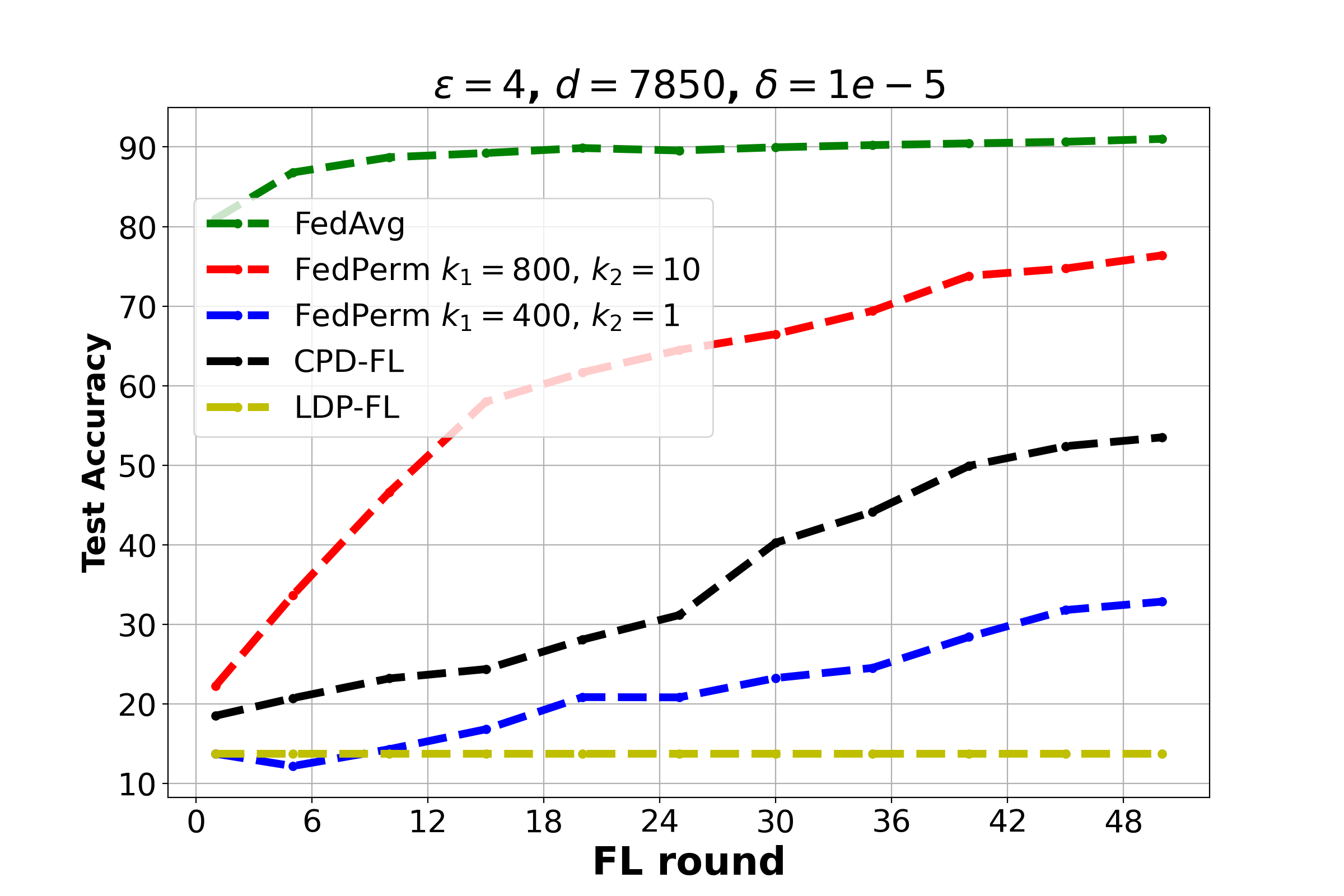} 
\caption{Accuracy per FL round.}
\label{fig:Acc2}
\end{subfigure}
\vspace{-0.1in}
\caption{Test accuracy for different FL algorithms for MNIST over 15 clients.}
\label{fig:AccAll}
\end{figure}

Figure~\ref{fig:Acc1} shows the test accuracy of that model that was
trained using different FL algorithms running for $T=50$ rounds.  The
MNIST dataset is divided across $n=15$ clients with a Dirichlet
distribution.  Figure~\ref{fig:Acc2} shows the test accuracy for these
algorithms per FL round when the total privacy budget is fixed to
$\varepsilon=4.0$.  We compared two versions of \Name in these
experiments: \textbf{(a)} \Name with $k_1=400$ and $k_2=1$ which is a
``light'' version where encryption and decryption time at clients
takes around $52.2$ and $2.4$ seconds respectively.  It also imposes
$21$ minutes computation time at the server. \textbf{(b)} \Name with
$k_1=800$ and $k_2=10$ which is a ``heavy'' version where client
encryption, decryption, and server aggregation time takes around
$32.1$ minutes,  $2.4$ seconds and $16.4$ hours
respectively.

As we mentioned earlier, \Name provides a trade-off between privacy
amplification and compute resources -- larger the values of $k_1$ and
$k_2$, greater are the compute resources for training, which in turn
provides higher privacy amplification that results in better model
utility.  The heavy version of \Name needs more resources to be as
fast as the lighter version, but it can provide much more utility
(because the privacy amplification is larger so the amount of noise is
added is smaller).  For instance, after $T=50$, and total privacy
budget $(4.0, 1e^{-5})$, the heavy version provides $72.38\%$ test
accuracy while the light version provides $32.85\%$ test accuracy.
From these figures we can see if we invest enough computation
resources in \Name, we can provide higher utility compared to CDP-FL,
without trusting the \Name server.  Non-private FedAvg, CDP-FL and
LDP-FL also provides $91.02\%$, $53.50\%$, and $13.74\%$ test
accuracies for the same total $(\varepsilon, \delta)$ respectively.

\subsection{Time Analysis}
\begin{figure}[h!]
\centering
\begin{subfigure}{.23\textwidth}
\centering
\includegraphics[width=.99\linewidth]{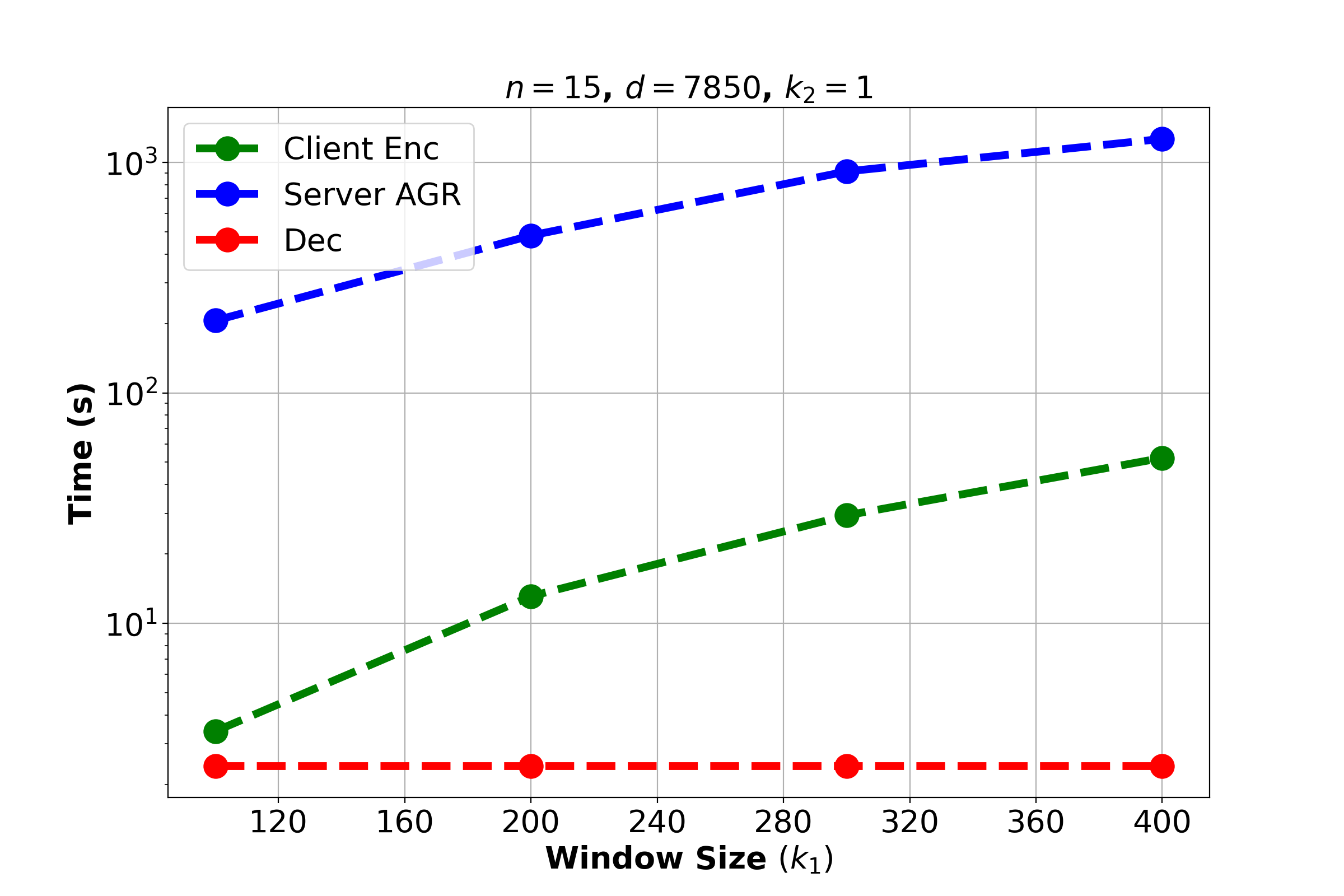} 
\caption{Impact of $k_1$.}
\label{fig:Time1}
\end{subfigure}
\begin{subfigure}{.23\textwidth}
\centering
\includegraphics[width=.99\linewidth]{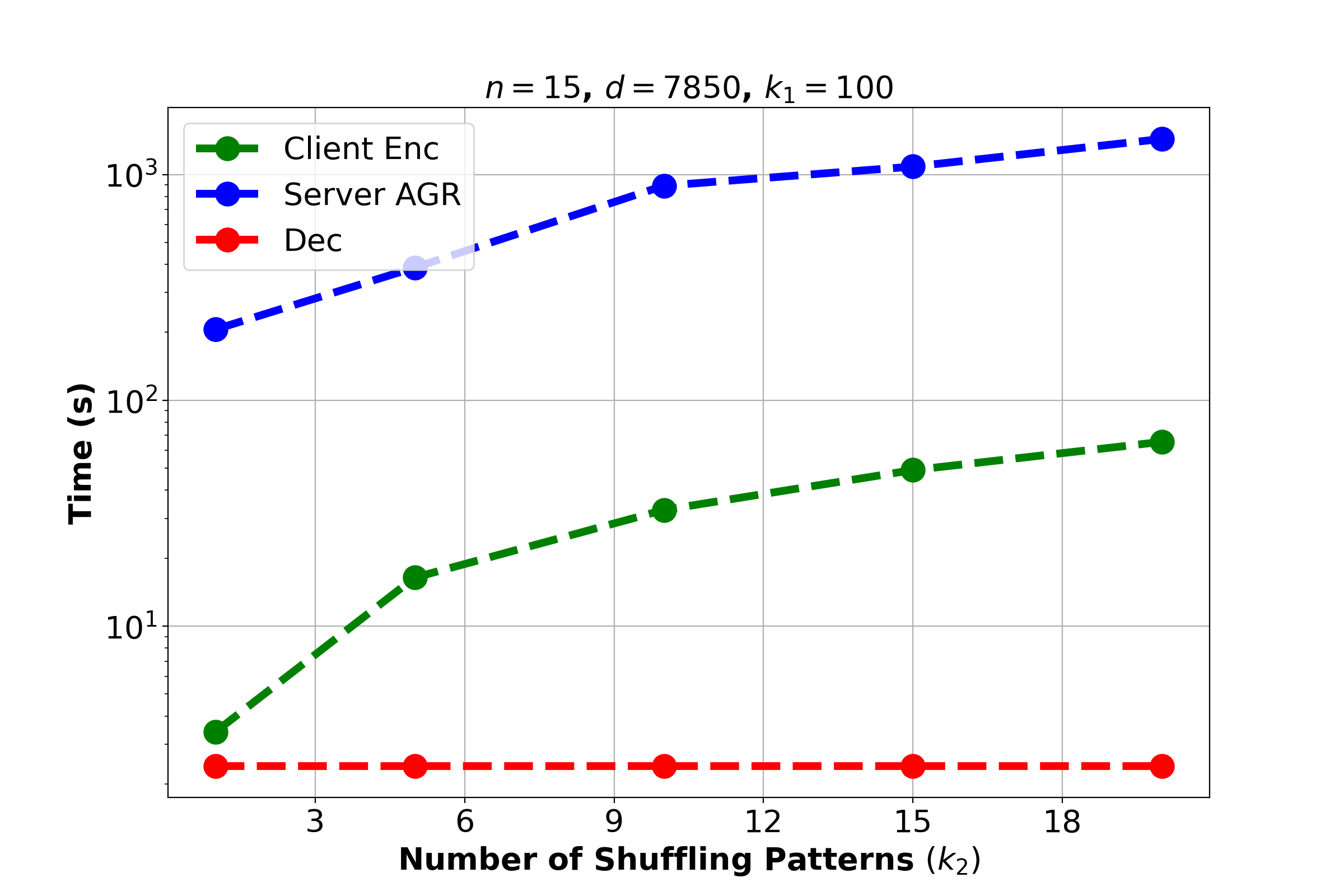} 
\caption{Impact of $k_2$.}
\label{fig:Time2}
\end{subfigure}
\begin{subfigure}{.23\textwidth}
\centering
\includegraphics[width=.99\linewidth]{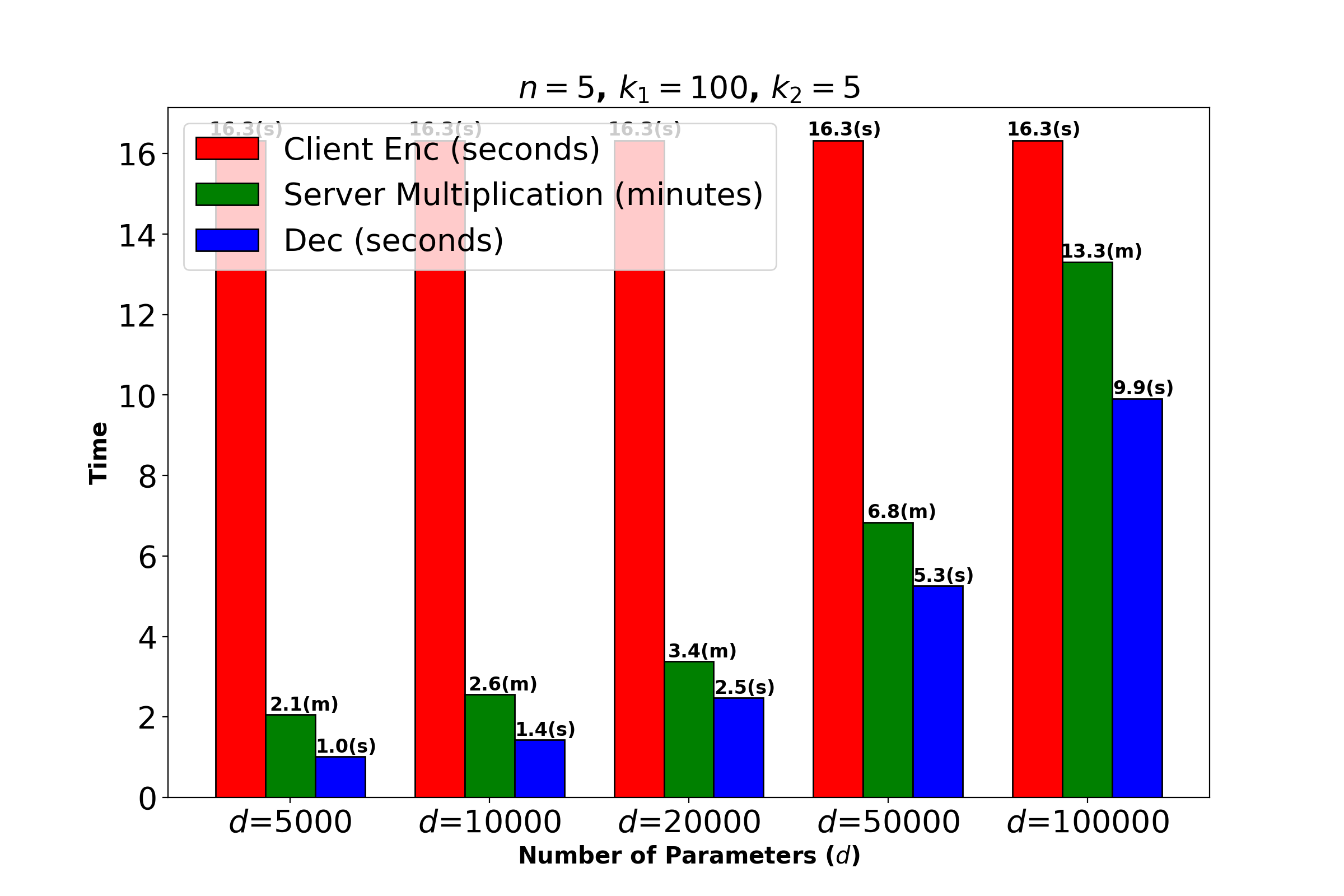} 
\caption{Impact of $d$.}
\label{fig:Time3}
\end{subfigure}
\begin{subfigure}{.23\textwidth}
\centering
\includegraphics[width=.99\linewidth]{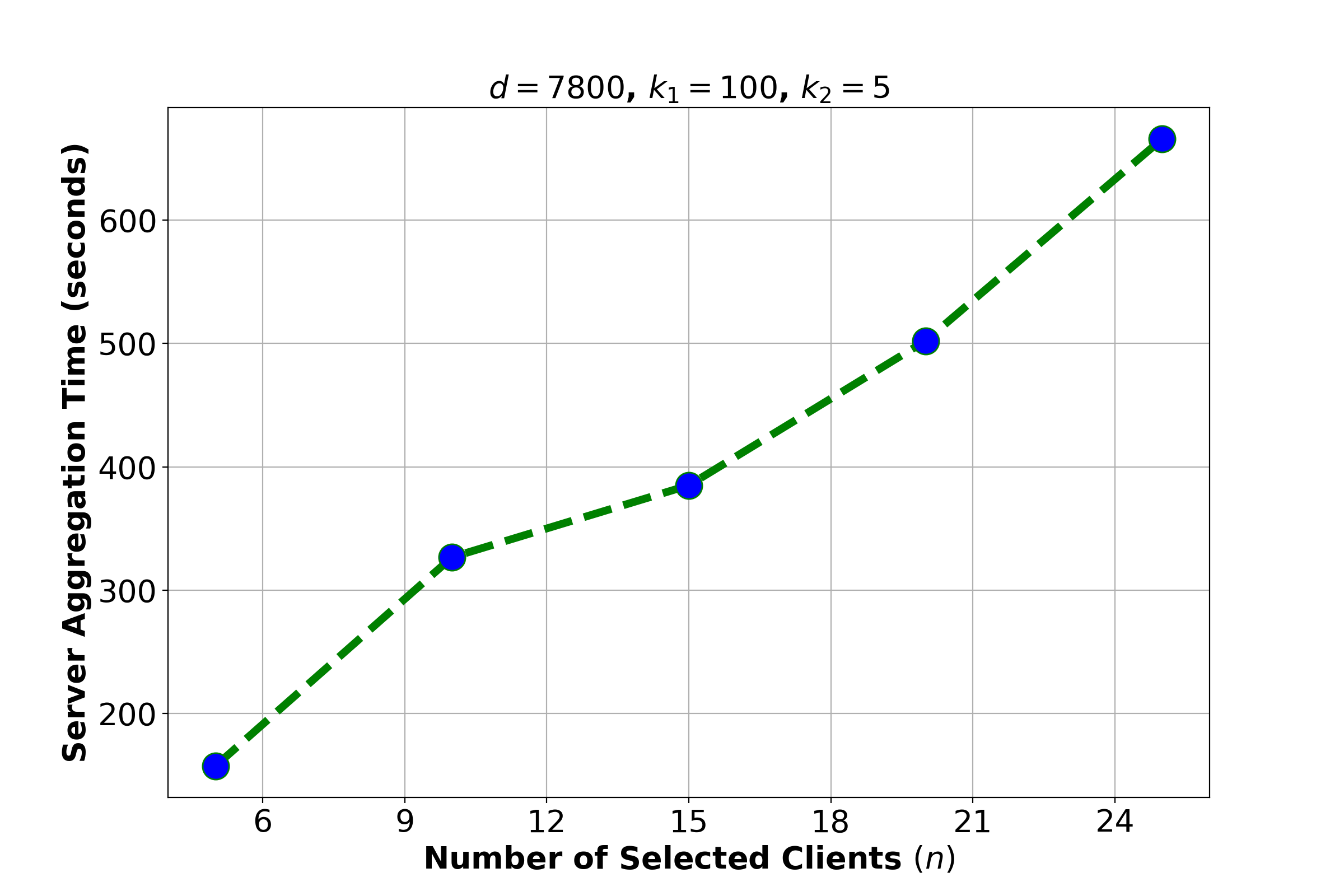} 
\caption{Impact of $n$.}
\label{fig:Time4}
\end{subfigure}
\vspace{-0.1in}
\caption{Client encryption, decryption, and server aggregation time in \Name .}
\label{fig:Time-ALL}
\end{figure}

We evaluate the impacts of our hyperparameters $k_1$, $k_2$, $n$, and
$d$s on the encryption, decryption and sever aggregation time in
Figure~\ref{fig:Time-ALL}.  We use Paillier encryption system and we
use a key size of $2048$ bits in our experiments.  For measuring time,
we use 64CPUs and 64GB memory for the client and server simulations.
Note that we opt to not use GPU as model training is not a bottleneck
in our system compared to HE operations.  Also note that these figures
are data independent as we are working with encryption and decryption
and homomorphic multiplication with plaintext and homomorphic
addition.

\paragraphb{Client encryption time:} In \Name, each client must do
$k_1^2 \cdot k_2$ encryptions for its query, therefore client
encryption time has a quadratic and linear relationship with window
size ($k_1$) and number of shuffling patterns ($k_2$) respectively
(Figures~\ref{fig:Time1} and~\ref{fig:Time2}). We also show in
Figure~\ref{fig:Priv-ALL} that increasing the $k_1$ has more impact
(close to quadratic impact) compared to increasing $k_2$ on the
privacy amplification. This means that if we invest more computation
resources on the clients and are able to do more encryption, we get
greater privacy amplification by parameter shuffling.  For instance,
when we increase the $k_1$ from $100$ to $200$ (while fixing $k_2=1$),
the average client encryption time increases from $3.4$ to $13.1$
seconds for $d=7850$ parameters.  And while fixing the $k_1=100$, if
we increase the number of shuffling patterns from $1$ to $10$, the
encryption time goes from $3.4$ to $32.7$ seconds.  When we fix the
value of $k_1$ and $k_2$, the number of encryption is fixed at the
clients, so the encryption time would be constant if we increase the
number of parameters ($d$) 
each round (Figure~\ref{fig:Time3}).

\paragraphb{Client decryption time:}
Changing $k_1$, $k_2$, and $n$ does not have any impact on decryption
time, as each client should decrypt $d$ parameters
(Figures~\ref{fig:Time1} and~\ref{fig:Time2}).  In
figure~\ref{fig:Time3}, we show the linear relationship of decryption
time and number of parameters.  For instance by increasing the number
of parameters from ${10}^5$ to ${10}^6$, the decryption time increases
from $1.01$ to $9.91$ seconds.

\paragraphb{Server aggregation time:} In \Name, the server first
multiplies the encrypted binary mask to the corresponding shuffled
model parameters for each client participating in the training round,
and then sums the encrypted unshuffled parameters to compute the
encrypted global model.  We employ joblib to parallelize matrix
multiplication over superwindows. Thus, larger the superwindows
greater is the parallelism. However, as we increase $k_1$ and/or $k_2$
the superwindow size goes down, and hence the parallelism, which leads
to increase in running time as observed in Figures~\ref{fig:Time1} and
~\ref{fig:Time2}.
Moreover, increasing $n$, $d$ has a linear relationship with server
aggregation time (Figure~\ref{fig:Time3} and~\ref{fig:Time4}).  For
instance, when we increase the $n$ from $5$ to $10$ the server
aggregation time increases from $157.47$ to $326.72$ seconds for
$d=7850$, $k_1=100$, and $k_2=1$.

\section{Conclusion}

We presented \Name, a new FL algorithm that combines LDP, intra-model
parameter shuffling at the federation clients, and a cPIR based
technique for parameter aggregation at the federation server to
deliver both client data privacy and robustness from model poisoning
attacks.  Our intra-model parameter shuffling significantly amplifies
the LDP guarantee for clients' training data.  The cPIR based
technique we employ allows cryptographic parameter aggregation at the
server.  At the same time, the server clips the clients' parameter
updates to ensure that model poisoning attacks by adversarial clients
are effectively thwarted.  We also presented windowing hyperparameters
in \Name that let us trade off compute resources with model utility.
Our empirical evaluation on the MNIST dataset demonstrates \Name's
privacy amplification benefits and studies the trade off between
computation and model utility.  We leave the study of extensions to
\Name\ -- (i) an additional dimension of the hyperparameters ($k_3$)
that takes the computation-utility trade offs to hypercube space (see
Appendix~\ref{sec:future-work}), (ii) plugging in other PIR protocols,
and (iii) combining an external client shuffler with \Name\ -- to
future work.

\bibliography{main} 

\appendix


\begin{figure}[h]
\centering
\includegraphics[width=0.9\columnwidth]{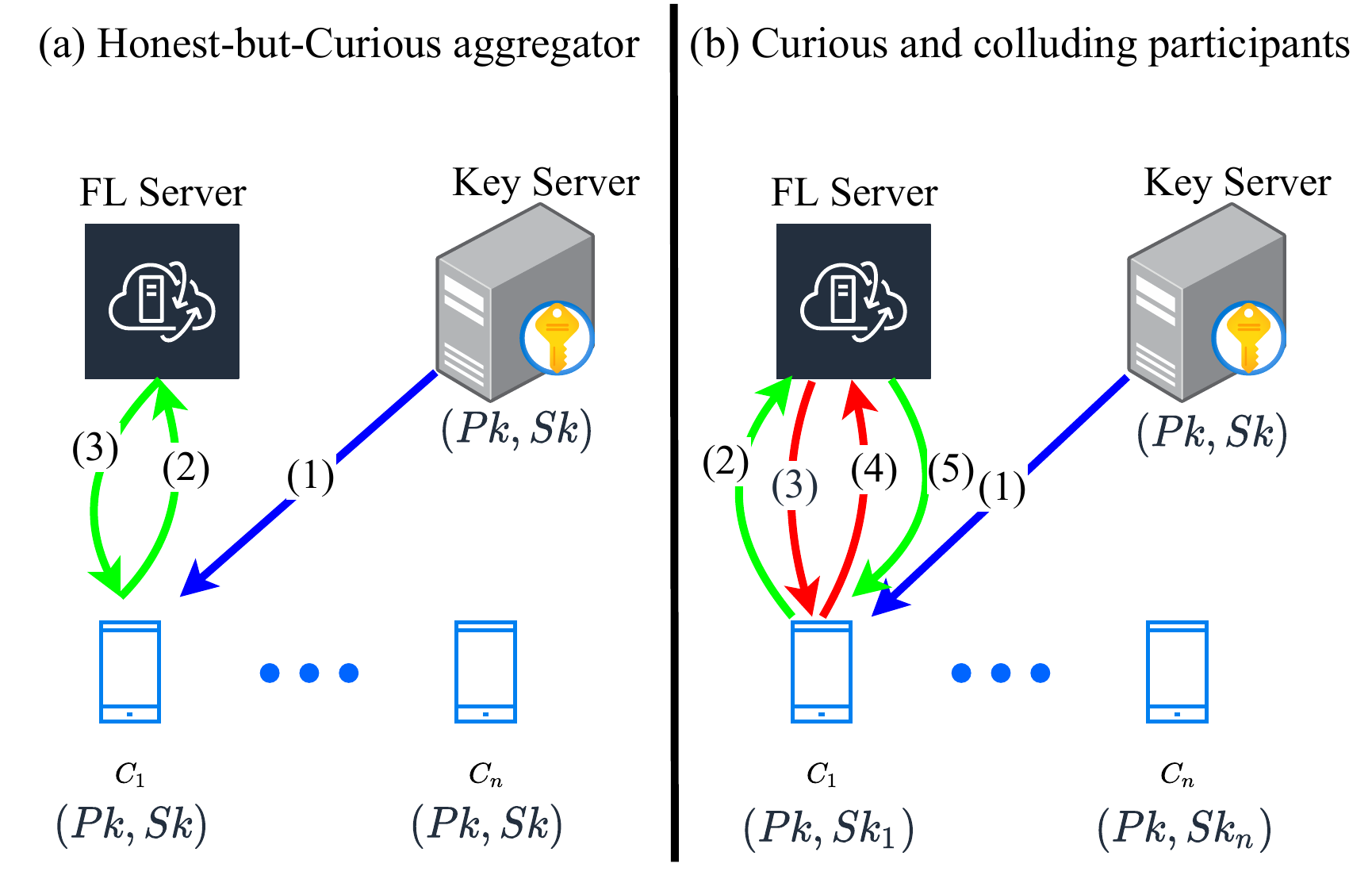} 
\caption{Different threat models.}
\label{fig:TM}
\end{figure}

\section{Threat Models}\label{Sec:TM}
In this section, we describe two threat models that are of interest to
our work, and illustrated in Figure~\ref{fig:TM}.

\subsection{Honest-but-Curious Aggregator}
In this threat model, we assume that the server correctly follows the
aggregation algorithm, but may try to learn clients' private
information by inspecting the model updates sent by the
participants. This is a common assumption that previous
works~\cite{zhang2020batchcrypt, xu2019hybridalpha,
  bonawitz2017practical, truex2019hybrid} also consider.  For creating
the PIR queries, we use Paillier~\cite{paillier1999public} homomorphic
encryption. We explain different homomorphic encryption systems that
we use in Appendix~\ref{sec:HE}.  Before starting \Name, we need a key
server to generate and distribute the keys for the homomorphic
encryption (HE).  A key server generates a pair of public and secret
homomorphic keys $(Pk, Sk)$, sends them to the clients, and sends only
the public key to the server.  Either a trusted external key server or
a leader client can be responsible for this role.  For the leader
client, similar to previous works~\cite{zhang2020batchcrypt}, before
the training starts, the FL server randomly selects a client as the
leader.  The leader client then generates the keys and distributes
them to the clients and the server as above.

The steps of \Name for this threat model (Figure~\ref{fig:TM}(a)) are
as follows:
\textbf{(1)} The pair of keys are distributed by the key server to all
the clients.
\textbf{(2)} In each round of training, the clients learn their local
updates, generate encrypted PIR query and shuffled parameters, and
send them to the server.  Next, the server aggregates the updates, and
sends the aggregated update to the clients.
\textbf{(3)} Each client can decrypt the encrypted global parameters
received from the server using the private key and updates its model.

\subsection{Curious and Colluding Clients}
In this threat model, we assume that some clients may collude with the
FL server to get private information about a victim client by
inspecting its model update.  For this threat model, we use
thresholded Paillier~\cite{damgaard2001generalisation}.  In the
thresholded Paillier scheme, the secret key is divided to multiple
shares, and each share is given to a different client.  For this
threat model, we need an external key server that generates the keys
and sends $(Pk, Sk_i)$ to each client, and sends the public key to the
server.  Now each client can partially decrypt an encrypted
message, but if less than a threshold, say $t$, combine their partial
decrypted values, they cannot get any information about the real
message.  On the other hand, if we combine $\geq t$ partial decrypted
values, we can recover the secret.  We explain how thresholeded
Paillier scheme works in Appendix~\ref{sec:HE}.

The steps of \Name for this threat model (Figure~\ref{fig:TM}(b)) are
as follows:
\textbf{(1)} The pairs of keys are distributed to the clients by the
key server.
\textbf{(2)} In each round of training, the clients learn their local
updates,generate encrypted PIR query and shuffled parameters, and send
them to the server.  Next, the server aggregates the local updates to
produce global model update (which is encrypted).
\textbf{(3)} The server randomly chooses $t$ clients to partially
decrypt the global model update. The \Name server sends the encrypted
global update to these clients.
\textbf{(4)} Each client decrypts the global model with its specific
partial secret key $Sk_i$, and sends the result back to the server.
\textbf{(5)} The server first authenticates each partial decryption
that is done by the true $Sk_i$ (by a zero-knowledge proof provided by
thresholded Paillier~\cite{damgaard2001generalisation}).  Then the
\Name server combines the partial decrypted updates and broadcasts
plain unshuffled model updates to all the clients for the next round
of \Name.

At present our implementation of \Name does not support this threat
model, and we leave it for future work.

\section{Central Differential Privacy in FL (CDP-FL)}
Algorithm~\ref{alg:CDP-FL} shows how CDP-FL works which is also
discussed in ~\cite{mcmahan2018learning, geyer2017differential,
  naseri2020local}.  In CDP-FL, the server receives model updates
capped by norm $C$, and after averaging them, it adds i.i.d sampled
noise to the parameters $\theta_g^{t+1} \gets \theta_g^{t} +
\frac{1}{n} \sum_{u \in U} \theta_u^{t} + \mathcal{N} (0, \sigma^2
\mathbb{I})$ where $\sigma \gets \frac{\Delta_2}{\varepsilon} \sqrt{2
  ln(1.25) / \delta }$ and the $\ell_2$ sensitivity is $\Delta_2=C$.

\begin{algorithm}[h!]
\caption{Central Differential Privacy in FL (CDP-FL)}
\label{alg:CDP-FL}
\textbf{Input}: number of FL rounds $T$, number of local epochs $E$,
number of all the clients $N$, number of selected users in each round
$n$, total privacy budget $TP$, probability of subsampling clients
$q$, learning rate $\eta$, noise scale $z$, bound $C$
\\ \textbf{Output}: global model $\theta_{g}^{T}$
\begin{algorithmic}[1] 
\STATE  $\theta_g^{0}  \gets$ Initialize weights
\STATE Initialize MomentAccountant$(\varepsilon,  \delta,  N)$

\FOR{each iteration $t \in [T]$}
	\STATE $U \gets$ set of $n$ randomly selected clients out of $N$ total clients with probability of $q$
	\STATE $p_t \gets $ MomentAccountant.getPrivacySpent()  \COMMENT{\% privacy budget spent till this round}
	\IF{$p_t > TP$} 
		\STATE \textbf{return} $\theta_g^{T}$ \COMMENT{\% if spent privacy budget is passed over the threshold finish FL training}
	\ENDIF
    	\FOR{$u$ in $U$}
		\STATE $\theta \gets \theta_g^{t}$
		\FOR{local eopoch $e \in [E]$}
			\FOR{batch $b \in [B]$}
				\STATE $\theta \gets \theta - \eta \triangledown L(\theta,  b)$ 
				\STATE $\triangle \gets \theta - \theta_g^{t}$
				\STATE $\theta \gets \theta_{g}^{t} + \triangle \min{(1,  \frac{C}{{||\triangle||}_2})}$
			\ENDFOR
		\ENDFOR
		\STATE Client $u$ sends $\theta_{u}^{t}=\theta-\theta_{g}^{t}$ to the server
	\ENDFOR
	\STATE $\sigma \gets z C/q$
	\STATE  $\theta_g^{t+1} \gets \theta_g^{t} +  \frac{1}{n} \sum_{u \in U} \theta_u^{t} + \mathcal{N} (0,  \sigma^2 \mathbb{I}) $
	\STATE MomentAccountant.accumulateSpentBudget$(z)$
\ENDFOR
\STATE \textbf{return} $\theta_g^{T}$
\end{algorithmic}
\end{algorithm}

\section{Laplace Mechanism}~\label{sec:Laplace}
The most common mechanism for achieving pure $\varepsilon_{\ell}$-DP
is Laplace mechanism~\cite{}, where

\begin{definition}
Let $f: \mathcal{X}^n \rightarrow \mathbb{R}^{k}$.  The
$\ell_1$-sensitivity of $f$ is:
\begin{equation}
\Delta_1^{(f)} = \max_{X, X'} {|| f(X) - f(X')||}_1
\end{equation}
where $X, X' \in \mathcal{X}^n$ are neighbering datasets differing
from each other by a single data record.
\end{definition}

Sensitivity gives an upper bound on how much the output of the
randomizer can change by switching over to a neighboring dataset as
the input.

\begin{definition}
Let $f: \mathcal{X}^n \rightarrow \mathbb{R}^{k}$. The Laplace
mechanism is defined as:
\begin{equation}
\mathcal{R} (X) = f(X) + [Y_1,  \dots, Y_k]
\end{equation}
Where the $Y_i$s are drawn i.i.d from
Laplace$(\Delta^{(f)}/\varepsilon)$ random variable.  This
distribution has density of
$p(x)=\frac{1}{2b}\exp{\left(-\frac{|x|}{b} \right)}$ where $b$ is the
scale and equal to $\Delta^{(f)}/\varepsilon$.
\end{definition}

In \Name, each client $i$ applies the Laplace mechanism as a
randomizer $\mathcal{R}$ on its local model update $(x_i)$.  Each
model update contains $d$ parameters in range of $[0,1]$, so the
sensitivity of the entire input vector is $d$.  It means that applying
$\varepsilon_d$-DP on the vector $x_i$ is equal to applying
$\varepsilon_{wd}=\varepsilon_{d}/d$ on each parameter independently.
Therefore, applying $\varepsilon_d$-DP randomizer $\mathcal{R}$ on the
vector $x_i$ means adding noise from Laplace distribution with scale
$b=\frac{1}{\varepsilon_{wd}}=\frac{1}{\frac{\varepsilon_d}{d}}=\frac{d}{\varepsilon_d}$.

\section{Background}
\subsection{Robustness to poisoning attacks} 
Most of the distributed learning algorithms, including FedAvg~\cite{mcmahan2017communication}, 
operate on mutually untrusted clients and server. This makes distributed learning susceptible to the threat of poisoning~\cite{kairouz2019advances, shejwalkar2021back}. 
A \emph{poisoning adversary} can either own or control a few of FL clients, called \emph{malicious clients}, and instruct them to share malicious updates with the central server in order to reduce the performance of the global model. There are two approaches to poisoning FL: \emph{untargeted}~\cite{baruch2019a, fang2020local,shejwalkar2021manipulating} attacks aim to reduce the utility of global model on arbitrary test inputs; and \emph{backdoor}~\cite{bagdasaryan2020backdoor,wang2020attack, xie2019dba} attacks aim to reduce the utility on test inputs that contain a specific signal called the trigger.

In order to make FL robust to  the presence of such malicious clients, the literature has designed various \emph{robust aggregation rules (AGR)}~\cite{blanchard2017machine, DBLP:conf/icml/MhamdiGR18, YinCRB18,mozaffari2021fsl}, which aim to remove or attenuate the updates that are more likely to be malicious according to some criterion. For instance, Multi-krum~\cite{blanchard2017machine} repeatedly removes updates that are far from the geometric median of all the updates, and Trimmed-mean~\cite{YinCRB18} removes the largest and smallest values of each update dimension and calculates the mean of the remaining values.
It is not possible to use these AGRs in secure aggregation as the parameters are encrypted.

\subsection{Private Information Retrieval (PIR)}
Private information retrieval (PIR) is a technique to provide query
privacy to users when fetching sensitive records from untrusted
database servers.  That is, PIR enables users to query and retrieve
specific records from untrusted database server(s) in a way that the
servers cannot identify the records retrieved.  There are two major
types of PIR protocols. The first type is \emph{computational PIR}
(CPIR)~\cite{chang2004single} in which the security of the protocol
relies on the computational difficulty of solving a mathematical
problem in polynomial time by the servers, e.g., factorization of
large numbers. Most of the CPIR protocols are designed to be run by a
single database server, and therefore to minimize privacy leakage they
perform their heavy computations on the whole database (even if a
single entry has been queried). Most of these protocols use
homomorphic encryption (Section~\ref{sec:HE}) to make their queries
private.  The second major class of PIR is \emph{information-theoretic
  PIR} (ITPIR)~\cite{mozaffari2020heterogeneous}.  ITPIR protocols
provide information-theoretic security, however, existing designs need
to be run on more than one database servers, and they need to assume
that the servers do not collude.  Our work uses computational PIR
(cPIR) protocols to make the shuffling private.

\subsection{Homomorphic Encryption (HE)}\label{sec:HE}
Homomorphic encryption (HE) allows application of certain arithmetic
operations (e.g., addition or multiplication) on the ciphertexts
without decrypting them.  Many recent works~\cite{chang2004single}
advocate using partial HE, that only supports addition, to make the FL
aggregation secure.  In this section we describe two important HE
systems that we use in our paper.

\paragraphb{Paillier}
An additively homomorphic encryption system provides following
property:

\begin{equation}
  Enc(m_1) \circ Enc(m_2) = Enc(m_1 + m_2)
\end{equation}
where $\circ$ is a defined function on top of the ciphertexts. 

In these works, the clients encrypt their updates, send them to the
server, then the server can calculate their sum (using the $\circ$
operation) and sends back the encrypted results to the clients.  Now,
the clients can decrypt the global model locally and update their
models.  Using HE in these scenario does not produce any accuracy loss
because no noise will be added to the model parameters during the
encryption and decryption process.

\paragraphb{Thresholded Paillier}
In~\cite{damgaard2001generalisation}, the authors extend the Paillier
system and proposed a thresholded version.  In the thresholded
Paillier scheme, the secret key is divided to multiple shares, and
each share is given to a different participant.  Now each participant
can partially decrypt an encrypted message, but if less than a
threshold, say $t$, combine their partial decrypted values, they
cannot get any information about the real message.  On the other hand,
if we combine $\geq t$ partial decrypted values, we can recover the
secret.  In this system, the computations are in group
$\mathbb{Z}_{n^2}$ where $n$ is an RSA modulus.  The process is as
follows:

\begin{itemize}
\item \emph{Key generation:} First find two primes $p$ and $q$
  that statisfies $p=2p'+1$ and $q=2q'+1$ where $p',q'$ are
  also prime.  Now set the $n=pq$ and $m=p'q'$.  Pick $d$ such
  that $d=0\; \text{mod} \; m$ and $d=1 \; \text{mod} \;
  n^2$. Now the key server creates a polynomial
  $f(x)=\sum_{i=0}^{k-1} a_i x^i \; \text{mod} \; n^2m$ where
  $a_i$ are chosen randomly from $\mathbb{Z}^{*}_{n^2m}$ and
  the secret is hidden at $a_0=d$.  Now each secret key share
  is calculated as $s_i=f(i)$ for $\ell$ shareholders and the
  public key is $n$. For verification of correctness of
  decryption another public value $v$ is also generated where
  the verification key for each shareholder is $v_i=v^{\Delta
    s_i} \; \text{mod} \; n^2$ and $\Delta=\ell$.
\item \emph{Encryption:} For message $M$, a random number $r$
  is chosen from $\mathbb{Z}^{*}_{n^2}$ and the output
  ciphertext is $c=g^{M} \cdot r^{n^2} \; \text{mod} \; n^2$.
\item \emph{Share decryption:} The $i^{th}$ shareholder
  computes $c_i = c^{2 \Delta s_i}$ for ciphertext $c$. And
  for zero-knowlege proof, it provides $\log_{c^4}
  {(c_i^2)}=\log_{v} {(v_i)}$ which provides gurantee that the
  shareholder really uses its secret share for decryption
  $s_i$.
\item \emph{Share combining:} After collecting $k$ partial
  decryption, the server can combine them into the original
  plain-text message $M$ by $c'=\Pi_{i\in [k]}
  c_i^{2\lambda_{0,i}^S} \; \text{mod} \; n^2$ where
  $\lambda_{0,i}^S=\Delta \Pi_{i' \in [k] , i'\ne i}
  \frac{-i}{i-i'}$.  And use it to generate the $M$.
\end{itemize}

\section{Discussion and Future Work}
\label{sec:future-work}
\paragraphb{Utilizing recursion in cPIR.}  A solution to reduce the
number encryptions and upload bandwidth at the clients would be using
recursion in our cPIR.  In this technique, the dataset is represented
by a $k_3$-dimensional hypercube, and this allows the PIR client to
prepare and send $k_3 \sqrt[k_3]{d}$ encrypted values where $k_3$
would be another hyperparameter.  For future work, we can use this
technique and reduce the number of encryptions which makes the upload
bandwidth consumption lower too.  For instance, if we have one
shuffling pattern $k_2=1$, the number of encryption decreases from
$k_1 d$ to $ k_3 \sqrt[k_3]{k_1} d$.

\paragraphb{Plugging newer PIR protocol.}  \Name utilizes cPIR for
private aggregation, and in particular we use ~\cite{chang2004single}
which is based on Paillier. However, any other cPIR protocol can be
used in \Name. For example, SealPIR~\cite{angel2018pir} can be used to
reduce the number of encryptions at the client. SealPIR is based on a
SEAL which is a lattice based fully HE. The authors show how to
compress the PIR queries and achieve size reduction of up to
$274\times$. We defer analyzing \Name with other cPIR schemes to
future work.

\paragraphb{Combination of an external client shuffler for more
  privacy amplification.}  For further privacy amplification, we can
use an external shuffler that shuffles the $n$ sampled clients'
updates similar to FLAME~\cite{liu2021flame}.  For future work, we can
use double amplification by first shuffling the parameters at the
clients (i.e. , detaching the parameter values from their position)
and then shuffle the client's updates at the external shuffler (i.e.,
detaching the updates from their client’s ID).

\end{document}